\pdfoutput=1

\documentclass[11pt]{article}

\usepackage{naacl2021}
\usepackage{times}
\usepackage{latexsym}

\usepackage[T1]{fontenc}

\usepackage[utf8]{inputenc} 

\usepackage{microtype}

\usepackage{url}
\usepackage{latexsym}
\usepackage{booktabs}       
\usepackage{amsfonts}       
\usepackage{nicefrac} 
\usepackage{latexsym}
\usepackage{multirow}
\usepackage{booktabs}
\usepackage{amsmath}
\usepackage{booktabs}
\usepackage{subfig}
\usepackage{xcolor}
\usepackage{colortbl}
\usepackage{graphicx}  
\usepackage{xcolor}
\usepackage{colortbl}
\usepackage[normalem]{ulem}
\usepackage[font={footnotesize}]{caption}

\usepackage{caption}
\usepackage{wrapfig}

\definecolor{myblue}{rgb}{0.9, 0.1, 0.94}
\definecolor{myyellow}{rgb}{0.98, 0.94, 0.75}
\definecolor{mypink}{rgb}{0.99, 0.87, 0.9}
\definecolor{mypink2}{rgb}{0.98, 0.38, 0.5}
\definecolor{brinkpink}{rgb}{0.98, 0.38, 0.5}

\definecolor{mygreen}{rgb}{0.0, 0.5, 0.0}
\definecolor{myorange}{rgb}{1.0, 0.49, 0.0}	
\definecolor{cadetgrey}{rgb}{0.86, 0.86, 0.86}
\definecolor{mygrey}{rgb}{0.9, 0.9, 0.9}
\definecolor{lightgray}{rgb}{0.83, 0.83, 0.83}
\definecolor{green}{rgb}{0.0, 0.5, 0.0}

\usepackage{microtype}

\title{Larger-Context Tagging: When and Why Does It Work?}

\author{Jinlan Fu$\dag$, Liangjing Feng$\dag$, Qi Zhang$\dag$, Xuanjing Huang$\dag$, Pengfei Liu$\sharp$ \thanks{\ \ Corresponding author}\\
$\dag$ School of Computer Science, Shanghai Key Laboratory \\
of Intelligent Information Processing, Fudan University,\\
  $\sharp$Carnegie Mellon University \\
  \texttt{\{fujl16,qz,xjhuang\}@fudan.edu.cn},  \texttt{pliu3@cs.cmu.edu}}

\begin{document}
\maketitle
\begin{abstract}
The development of neural networks and pretraining techniques has spawned many sentence-level tagging systems that achieved superior performance on typical benchmarks. However, a relatively less discussed topic is what if more context information is introduced into current top-scoring tagging systems. Although several existing works have attempted to shift tagging systems from sentence-level to document-level, there is still no consensus conclusion about when and why it works, which limits the applicability of the larger-context approach in tagging tasks. In this paper, instead of pursuing a state-of-the-art tagging system by architectural exploration, we focus on investigating when and why the larger-context training, as a general strategy, can work.

To this end, we conduct a thorough comparative study on four proposed aggregators for context information collecting and present an attribute-aided evaluation method to interpret the improvement brought by larger-context training. Experimentally, we set up a testbed based on four tagging tasks and thirteen datasets. Hopefully, our preliminary observations can deepen the understanding of larger-context training and enlighten more follow-up works on the use of contextual information.
We have released all relevant codes for future researchers to run similar analyses:
\url{https://github.com/jlfu/larger-context}.
\end{abstract}

\section{Introduction} \label{sec:intro}

The rapid development of deep neural models has shown impressive performances on sequence tagging tasks that aim to assign labels to each token of an input sequence \cite{sang2003introduction,lample2016neural,ma2016end}. More recently, the use of unsupervised pre-trained models \cite{akbik2018contextual,akbik2019pooled,peters2018deep,devlin2018bert} (especially contextualized version) has driven state-of-the-art performance to a new level.
Among these works, researchers frequently choose the boundary with the granularity of sentences for tagging tasks (i.e., \emph{sentence-level tagging}) \cite{huang2015bidirectional,chiu2015named,ma2016end,lample2016neural}.
Undoubtedly, as a transient, sentence-level setting enables us to develop numerous successful tagging systems, nevertheless the task itself should have not be defined as sentence-level but for simplifying the learning process for machine learning models.
Naturally, it would be interesting to see what if larger-context information (e.g., taking information of neighbor sentences into account) is introduced to modern top-scoring systems, which have shown superior performance under the sentence-level setting.
A small number of works have made seminal exploration in this direction, in which part of works show significant improvement of larger-context~\cite{luo2020hierarchical,xu2019document}
while others don't \cite{hu2020leveraging,hu2019document,luo2018attention}. 
Therefore, it's still unclear when and why larger-context training is beneficial for tagging tasks.
In this paper, we try to figure it out by asking the following three research questions:

\noindent \textbf{Q1}: \textit{How do different integration ways of larger-context information influence the system's performance?}
The rapid development of neural networks provides us with diverse flavors of neural components to aggregate larger-context information, which, for example, can be structured as a sequential topology by \emph{recurrent neural networks}~\cite{ma2016end,lample2016neural} (RNNs) 
or graph topology by \emph{graph neural networks}~\cite{kipf2016semi,schlichtkrull2018modeling}. 

Understanding the discrepancies of these aggregators can help us reach a more generalized conclusion about the effectiveness of larger-context training.
To this end, we study larger-context aggregators with three different structural priors  (defined in Sec.~\ref{sec:aggregator}) and comprehensively evaluate their efficacy.

\noindent \textbf{Q2}: \textit{Can the larger-context training easily play to its strengths with the help of recently arising contextualized pre-trained models ~\cite{akbik2018contextual,akbik2019pooled,peters2018deep,devlin2018bert} (e.g. BERT)?}
The contextual modeling power of these pre-trained methods makes it worth looking at its effect on larger-context training.
In this work, we take BERT as a case study and assess its effectiveness quantitatively and qualitatively.

\noindent \textbf{Q3}: \textit{If improvements could be observed, where does the gain come and how do different characteristics of datasets affect the amount of gain?}
Instead of simply figuring out whether larger-context training could work, we also try to interpret its gains. Specifically, we propose to use fine-grained evaluation to explain where the improvement comes from and why different datasets exhibit discrepant gains.

Overall, the first two questions aim to explore \emph{when} larger-context training can work while the third question addresses  \emph{why}. 
Experimentally, we try to answer these questions by conducting a  comprehensive analysis, which involves four tagging tasks and thirteen datasets.
Our main observations are summarized in Sec.~\ref{sec:discussion}.~\footnote{Putting the conclusion at the end can help the reader understand it better since more contextual information about experiments has been introduced.}
Furthermore, we show, with the help of these observations, it's easier to adapt larger-context training to modern top-performing tagging systems with significant gains.
We brief our contributions below:

1) We try to bridge the gap by asking three research questions, between the increasing top-performing sentence-level tagging systems and insufficient understanding of larger-context training, encouraging future research to explore more larger-context tagging systems. 
2) We systematically investigate four aggregators for larger-context and present an attribute-aided evaluation methodology to interpret the relative advantages of them, and why they can work (Sec.~\ref{sec:aggregator}). 
3) Based on some of our observations, we adapt larger-context training to five modern top-scoring systems in the NER task and observe that all larger-context enhanced models can achieve significant improvement (Sec.~\ref{sec:top-systems}).
Encouragingly , with the help of larger-context training, the performance of \newcite{akbik2018contextual} on the \texttt{WB} (OntoNotes5.0-WB) dataset can be improved by a \textbf{10.78} $F1$ score.

\section{Task, Dataset, and Model}
We first explicate the definition of tagging task and then describe several popular datasets as well as typical methods of this task.

\subsection{Task Definition}
Sequence tagging aims to assign one of the pre-defined labels to each token in a sequence. 
In this paper, we consider four types of concrete tasks: Named Entity Recognition (NER),  Chinese Word Segmentation (CWS), Part-of-Speech (POS) tagging, and Chunking.

\subsection{Datasets}
The datasets used in our paper are naturally ordered without random shuffling according to the paper that constructed these datasets, except for WNUT-2016 dataset.

\noindent \textbf{Named Entity Recognition (NER)} We consider two well-established benchmarks: CoNLL-2003 (\texttt{CN03}) and OntoNotes 5.0. OntoNotes 5.0 is collected from six different genres: broadcast conversation (\texttt{BC}), broadcast news (\texttt{BN}), magazine (\texttt{MZ}), newswire (\texttt{NW}), telephone conversation (\texttt{TC}), and web data (\texttt{WB}). 
Since each domain of OntoNotes 5.0 has its nature, we follow previous works \cite{durrett2014joint,chiu2016named,ghaddar2018robust} that utilize different domains of this dataset, which also paves the way for our fine-grained analysis.

\noindent \textbf{Chinese Word Segmentation (CWS)} We use four mainstream datasets
from SIGHAN2005 and SIGHAN2008, in which
\texttt{CITYU} is traditional Chinese, while \texttt{PKU},  \texttt{NCC}, and \texttt{SXU} are simplified ones.

\noindent \textbf{Chunking (Chunk)} CoNLL-2000 (\texttt{CN00}) is a benchmark dataset for text chunking.

\noindent \textbf{Part-of-Speech (POS)} We use the Penn Treebank (\texttt{PTB}) III dataset for POS tagging.\footnote{It's hard to cover all datasets for all tasks. For Chunk and POS tasks, we adopt the two most popular benchmark datasets.}    

\subsection{Neural Tagging Models} \label{sec:taggmodels}
Despite the emergence of a bunch of architectural explorations \cite{ma2016end,lample2016neural,yang2018design,peters2018deep,akbik2018contextual,devlin2018bert} for sequence tagging,
two general frameworks can be summarized:
(i) \textit{cEnc-wEnc-CRF} consists of the word-level encoder, sentence-level encoder, and CRF layer \cite{lafferty2001conditional}; 
(ii) \textit{ContPre-MLP} is composed of a contextualized pre-trained layer, followed by an MLP or CRF layer.
In this paper, we take both frameworks as study objects for our three research questions first, \footnote{Notably, in the setting, we don't aim to improve performance over state-of-the-art models.} and instantiate them as two specific models: \textit{CNN-LSTM-CRF} \cite{ma2016end}  and \textit{BERT-MLP} \cite{devlin2018bert}.

\section{Larger-Context Tagging}
\subsection{Sentence-level Tagging}
Let ${S} = s_{1}, \cdots, s_{k}$ represent a sequence of sentences, where sentence $s_{i}$ contains $n_i$ words: $s_{i} = w_{i,1}, \cdots, w_{i,n_{i}}$.
Sentence-level tagging models predict the label for each word $w_{i,t}$ sentence-wisely (within a given sentence $s_i$).
\textit{CNN-LSTM-CRF}, for example, first converts
each word $w_{i,t} \in s_i$ into a vector by different word-level encoders $\textrm{wEnc}(\cdot)$:

{\small
\begin{align}
    \mathbf{w}_{i,t} &= \mathrm{wEnc}(w_{i,t}) = \mathrm{Lookup}(w_{i,t}) \oplus \mathrm{CNN}(w_{i,t}),  \label{eq:wenc} 
\end{align}
}

\noindent
where $\oplus$ denotes the concatenation operation, $ \mathrm{Lookup}(w_{i,t})$ can be pre-trained by context-free (e.g., GloVe) or context-dependent (e.g., BERT) word representations.

And then the concatenated representation of them will be fed into sentence encoder $\mathrm{sEnc}(\cdot)$ (e.g., LSTM layer) to derive a contextualized representation for each word.

\begin{align}
    \mathbf{h}_{i,t} &= \mathrm{sEnc}(\cdot) = \mathrm{LSTM}^{(s)}(\mathbf{w}_{i,t}, \mathbf{h}_{i,t-1}, \theta), \label{eq:basic_lstm}
\end{align}

\noindent
where the lower case ``$s$'' of $\mathrm{LSTM}^{(s)}$ represents a sentence-level LSTM.
Finally,  a CRF layer will be used to predict the label for each word.

\subsection{Contextual Information Aggregators}
\label{sec:aggregator}
\begin{figure*}%
\centering
\includegraphics[width=0.95\linewidth]{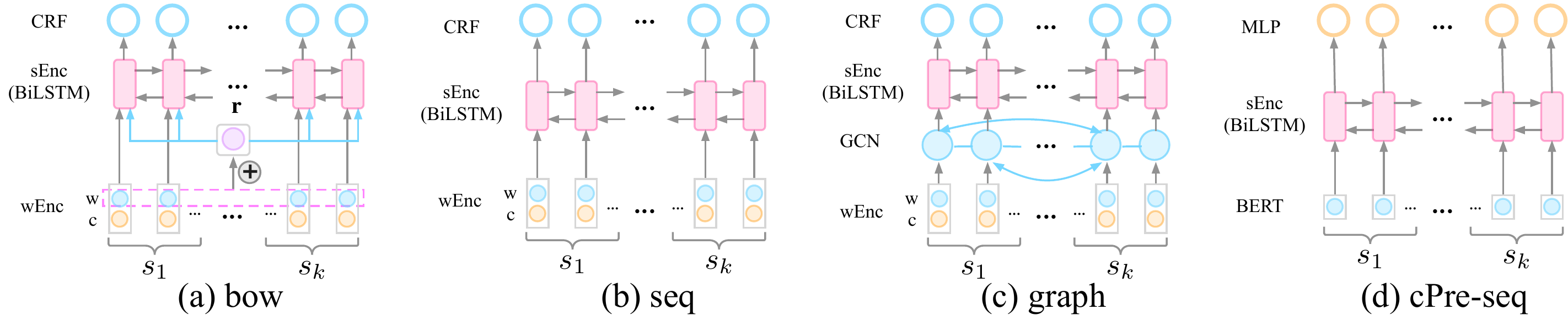}
\caption{Illustration of four larger-context aggregators.
} 
\label{fig:for_models}
\end{figure*}

Instead of predicting entity tags sentence-wisely, more contextual information of neighbor sentences can be introduced in diverse ways.
Following, we elaborate on how to extend sentence-level tagging to a larger-context setting.
The high-level idea is to introduce more contextual information into word- or sentence-level encoder defined in Eq.~\ref{eq:wenc} and Eq.~\ref{eq:basic_lstm}. Here, we propose four larger-context aggregators, whose architectures are illustrated in Fig.~\ref{fig:for_models}.

\paragraph{Bag-of-Word Aggregator (\textit{bow})} 
calculates a fused representation $\mathbf{r}$ for a sequence of sentences.

\begin{align}
        \mathbf{r} &= \mathrm{BOW}({w}_{1,1}, \cdots,{w}_{1,n_1}, \cdots,{w}_{k,n_k}),
\end{align}

\noindent
where $\mathrm{BOW}(\cdot)$ is a function that computes the average of all word representations of input sentences.
Afterward, $\mathbf{r}$, as additional information, will be injected into the word encoder.

More precisely, the word-level encoder  and sentence-level encoder can be re-written below:

\begin{align}
    \mathbf{w}^{bow}_{i,t} &=\mathrm{GloVe}(w_{i,t})\oplus \mathrm{CNN}(w_{i,t})\oplus \mathbf{r}, \\
    \mathbf{h}^{bow}_{i,t} &= \mathrm{LSTM}^{(S)}(\mathbf{w}^{bow}_{i,t}, \mathbf{h}^{bow}_{i,t-1}, \theta), \label{eq:LSTM-bow} 
 \end{align}

\noindent
where the upper case ``S'' of $\mathrm{LSTM}^{(S)}$ denotes the larger-context encoder that utilizes an LSTM deal with a sequence of sentences ($S= s_1, \cdots, s_k$) (instead of solely one sentence).

\paragraph{Sequential Aggregator (\textit{seq})} 
first concatenates all sentences $s_i \in S$ and then encode it with a larger-context encoder $\mathrm{LSTM}^{(S)}$. Formally, \textit{seq} aggregator can be represented as:

\begin{align}
    \mathbf{h}^{seq}_{i,t} &= \mathrm{LSTM}^{(S)}(\mathbf{w}^{seq}_{i,t}, \mathbf{h}^{seq}_{i,t-1}, \theta), \label{eq:LSTM-seq} 
\end{align}

\noindent
where $\mathbf{w}^{seq}_{i,t}$ is defined as Eq.~\ref{eq:wenc}, and the $ \mathrm{Lookup}(w_{i,t})$ is GloVe.
Then, a CRF decoder is utilized to predict the tags for each word.

\paragraph{Graph Aggregator  (\textit{graph})} 
incorporates \emph{non-local} bias into tagging models. 
Each word $w_i$ is conceptualized as a node. For edge connections, we define the following types of edges between
pairs of nodes (i.e. $w_i$ and $w_j$) to encode various structural information in the context graph:
i) if $|i-j|=1$;
ii) if $w_i = w_j$. 
In practice, the \textit{graph} aggregator first collects contextual information over a sequence of sentences, and generate the word representation:

\begin{align}
    \mathbf{G} &= \mathrm{GraphNN}(\mathbf{V}, E, \theta), 
\end{align}

\noindent
where $\mathbf{V} = \{\mathbf{w}_{1,1}, \cdots, \mathbf{w}_{1,n_1},\cdots, \mathbf{w}_{k,n_k}\}$ and $\mathbf{w}_i$ can be obtained as defined in Eq.~\ref{eq:wenc}.
Additionally, 
$\mathbf{G} = \{\mathbf{g}_{1,1}, \cdots, \mathbf{g}_{1,n_1},\cdots, \mathbf{g}_{k,n_k}\}$
stores aggregated contextual information for each word. We instantiate $\mathrm{GraphNN}(\cdot)$ as \emph{graph convolutional neural networks} \cite{kipf2016semi}.

Afterwords, the contextual vector $\mathbf{g}$ will be introduced into larger-context encoder , i.e., $\mathrm{LSTM}^{(S)}$:

\begin{align}
     \mathbf{h}^{graph}_{i,t} &= \mathrm{LSTM}^{(S)}(\mathbf{g}_{i,t}, \mathbf{h}^{graph}_{i,t-1}, \theta), \label{eq:LSTM-graph} 
\end{align}

\paragraph{Contextualized Sequential Aggregator (\textit{cPre-seq})} is 
an extension of \textit{seq} aggregator by using contextualized pre-trained models, such as BERT~\cite{devlin2018bert}, Flair~\cite{akbik2018contextual}, and ELMo~\cite{peters2018deep}, as a word encoder. Here, \textit{cPre-seq} is instantiated as BERT to get the word representation, then followed by a larger-context encoder $\mathrm{LSTM}^{(S)}$. We make the length of larger-context for the \textit{cPre-seq} aggregator within \textbf{512}. \textit{cPre-seq} can be formalized as:

\begin{align}
    \mathbf{h}^{cPre}_{i,t} &= \mathrm{LSTM}^{(S)}(\mathrm{BERT}(w_{i,t}), \mathbf{h}^{cPre}_{i,t-1}, \theta). \label{eq:LSTM-seq-bert} 
\end{align}

\section{Experiment: When Does It Work?}
The experiment in this section is designed to answer the first two research questions: \textbf{Q1} and \textbf{Q2} (Sec.~\ref{sec:intro}).
Specifically, we investigate whether larger-context training can achieve improvement and how different structures of aggregator, contextualized pre-trained models influence it.

\paragraph{Settings and Hyper-parameters}
We adopt \textit{CNN-LSTM-CRF} as a prototype and augment it with larger-context information by four categories of aggregators: \textit{bow}, \textit{seq},  \textit{graph}, and \textit{cPre-seq}. 
We use Word2Vec~\cite{mikolov2013distributed} (trained on simplified Chinese Wikipedia dump) as non-contextualized embeddings for CWS task, and  GloVe~\cite{pennington2014glove}  for NER, Chunk, and POS tasks.

The window size (the number of sentence) $k$ of larger-context aggregators will be explored with a range of $k=\{1,2,3,4,5,6,10\}$ for \textit{seq},  \textit{bow}, and \textit{cPre-seq}.
We chose the best performance that the larger-context aggregator achieved with window size $k \neq 1$ as the final performance of a larger-context aggregator.~\footnote{The settings of window size $k$ are listed in the appendix. }
We use the result from the model with the best validation set performance, terminating training  when the performance on development is not improved in 20 epochs.

For the POS task, we adopt dataset-level accuracy as evaluated metric while for other tasks, we use a corpus-level $F1$-score \cite{sang2003introduction} to evaluate.

\renewcommand\tabcolsep{2pt}
\begin{table*}[!htb]
  \centering \footnotesize
    \begin{tabular}{clccccccccccccccc}
    \toprule
   \multirow{2}[3]{*}{\textbf{Emb.}} &\multirow{2}[3]{*}{\textbf{Agg.}} & \multicolumn{4}{c}{\textbf{CWS}}              & \multicolumn{7}{c}{\textbf{NER}}                      & \textbf{Chunk} & \textbf{POS} & \multirow{2}[3]{*}{\textbf{Avg.}} & \textbf{Signi.} \\
\cmidrule(lr){3-6}\cmidrule(lr){7-13}\cmidrule(lr){14-14}\cmidrule(lr){15-15}      & & \texttt{CITYU} & \texttt{NCC} & \texttt{SXU} & \texttt{PKU} & \texttt{CN03} & \texttt{BC} & \texttt{BN} & \texttt{MZ} & \texttt{WB} & \texttt{NW} & \texttt{TC} & \texttt{CN00} & \texttt{PTB} &       & \tiny{($\times10^{-2}$)}  \\
\midrule
         & \textit{norm}   & \multicolumn{1}{>{\columncolor{mygrey}}c}{93.70}  & \multicolumn{1}{>{\columncolor{mygrey}}c}{92.26}  & \multicolumn{1}{>{\columncolor{mygrey}}c}{94.94}  & \multicolumn{1}{>{\columncolor{mygrey}}c}{94.35}  & 90.46  & 75.38  & 86.89  & 85.42  & 62.09  & 88.38  & 63.69  & \multicolumn{1}{>{\columncolor{mygrey}}c}{93.85}  & 97.25  & \multicolumn{1}{>{\columncolor{mygrey}}c}{-}  & -  \\
    \textbf{\textbf{Non-}} & \textit{bow}  & \multicolumn{1}{>{\columncolor{mygrey}}c}{+0.17}  & \multicolumn{1}{>{\columncolor{mygrey}}c}{+0.42}  & \multicolumn{1}{>{\columncolor{mygrey}}c}{+0.03}  & \multicolumn{1}{>{\columncolor{mygrey}}c}{+0.04}  & \textcolor{brinkpink}{-0.39}  & +1.66  & +0.32  & +1.51  & +3.49  & +0.92  & +0.42  & \multicolumn{1}{>{\columncolor{mygrey}}c}{\textcolor{brinkpink}{-0.29}}  & \textcolor{brinkpink}{-0.14}  & \multicolumn{1}{>{\columncolor{mygrey}}c}{+0.54}  & 1.74  \\
    \textbf{\textbf{Con.}} & \textit{graph}  & \multicolumn{1}{>{\columncolor{mygrey}}c}{\textcolor{brinkpink}{-0.15}}  & \multicolumn{1}{>{\columncolor{mygrey}}c}{\textcolor{brinkpink}{-0.61}}  & \multicolumn{1}{>{\columncolor{mygrey}}c}{\textcolor{brinkpink}{-0.02}}  & \multicolumn{1}{>{\columncolor{mygrey}}c}{+0.33}  & +1.47  & +0.17  & +0.42  & \textcolor{brinkpink}{-0.16}  & +4.84  & +0.34  & +0.90  & \multicolumn{1}{>{\columncolor{mygrey}}c}{\textcolor{brinkpink}{-0.15}}  & +0.17  & \multicolumn{1}{>{\columncolor{mygrey}}c}{+0.61}  & 2.17  \\
      &     \textit{seq}   & \multicolumn{1}{>{\columncolor{mygrey}}c}{+0.27}  & \multicolumn{1}{>{\columncolor{mygrey}}c}{+0.34}  & \multicolumn{1}{>{\columncolor{mygrey}}c}{+0.18}  & \multicolumn{1}{>{\columncolor{mygrey}}c}{+0.08}  & \textcolor{brinkpink}{-0.14}  & +0.65  & \textcolor{brinkpink}{-0.50}  & +1.49  & +5.61  & +1.13  & +2.39  & \multicolumn{1}{>{\columncolor{mygrey}}c}{\textcolor{brinkpink}{-0.08}}  & +0.03  & \multicolumn{1}{>{\columncolor{mygrey}}c}{+0.77}  & 0.86  \\
    \midrule
    \multirow{2}[2]{*}{\textbf{Con.}} & \textit{norm}  & \multicolumn{1}{>{\columncolor{mygrey}}c}{97.09}  & \multicolumn{1}{>{\columncolor{mygrey}}c}{95.77}  & \multicolumn{1}{>{\columncolor{mygrey}}c}{97.49}  & \multicolumn{1}{>{\columncolor{mygrey}}c}{96.47}  & 90.77  & 80.46  & 89.67  & 87.03  & 68.78  & 90.04  & 63.34  & \multicolumn{1}{>{\columncolor{mygrey}}c}{96.45}  & 97.62  & \multicolumn{1}{>{\columncolor{mygrey}}c}{-}  & -  \\
      &    \textit{cPre}  & \multicolumn{1}{>{\columncolor{mygrey}}c}{+0.07}  & \multicolumn{1}{>{\columncolor{mygrey}}c}{+0.07}  & \multicolumn{1}{>{\columncolor{mygrey}}c}{+0.13}  & \multicolumn{1}{>{\columncolor{mygrey}}c}{+0.14}  & +0.72  & +1.27  & +0.39  & +0.19  & +7.26  & +0.99  & +6.00  & \multicolumn{1}{>{\columncolor{mygrey}}c}{+0.11}  & +0.04  & \multicolumn{1}{>{\columncolor{mygrey}}c}{+1.15}  & 0.26  \\
    \bottomrule
    \end{tabular}%
  \caption{
    The relative improvement (the performance difference between a model with larger-context aggregator (e.g. \textit{bow}) and the one without it) on tasks {CWS}, {NER}, {Chunk}, and POS. 
    ``\textbf{norm}'' denotes the normal setting ($K = 1$).
 The values in red are the performance of larger-context tagging ($k>1$) lower than sentence-level tagging ($k=1$). \textit{``Signi.''} denotes p-value of  ``significant test''. 
 ``\textit{Emb.}'', ``\textit{Non-Con.}'', ``\textit{Con.}'', and ``\textit{Agg.}'' are the abbreviations of ``\textit{Embeddings}'', ``\textit{Non-Contextualized}'',  ``\textit{Contextualized}'', and ``\textit{Aggregator}'' respectively.
 The values in pink indicate that the value is less than zero.
  }
  \label{tab:holist-norm-large}%
\end{table*}%

\subsection{Exp-I: Effect of Structured Typologies}
\label{sec:structured-effect}
Tab.~\ref{tab:holist-norm-large} illustrates the relative improvement results of four larger-context training ($k>1$) relative to the sentence-level tagging ($k=1$).
To examine whether the larger-context aggregation method has a significant improvement over the sentence-level tagging, we used significant test with Wilcoxon Signed-RankTest~\cite{wilcoxon1970critical} at $p=0.05$ level. 
Results are shown in Tab.~\ref{tab:holist-norm-large} (the last column). We find that improvements brought by four larger-context aggregators are statistically significant ($p<0.05$), suggesting that the introduction of larger-context can significantly improve the performance of sentence-level models.

\paragraph{Results} We detail main observations in Tab.~\ref{tab:holist-norm-large}:

\noindent 1) 
For most of the datasets, introducing larger-context information will bring gains regardless of the ways how to introduce it (e.g. \textit{bow} or \textit{graph}), indicating the efficacy of larger contextual information. Impressively, the performance on dataset \texttt{WB} is significantly improved by \textbf{7.26} F1 score with the \textit{cPre-seq} aggregator ($p=5.1\times10^{-3}<0.05$).

\noindent 2)
Overall, comparing with \textit{bow} and \textit{graph} aggregators, \textit{seq} aggregator has achieved larger improvement by average, which can be further enhanced by introducing contextualized pre-trained models (e.g. BERT).

\noindent 3)
Incorporating larger-context information with some aggregators also can lead to performance drop on some datasets (e.g, using \textit{graph} aggregator on dataset \texttt{MZ} lead to 0.16 performance drop), which suggests the importance of a better match between datasets and aggregators.

\begin{figure*}[!ht]
    \centering \footnotesize
     \subfloat[CWS]{
    \includegraphics[width=0.375\linewidth]{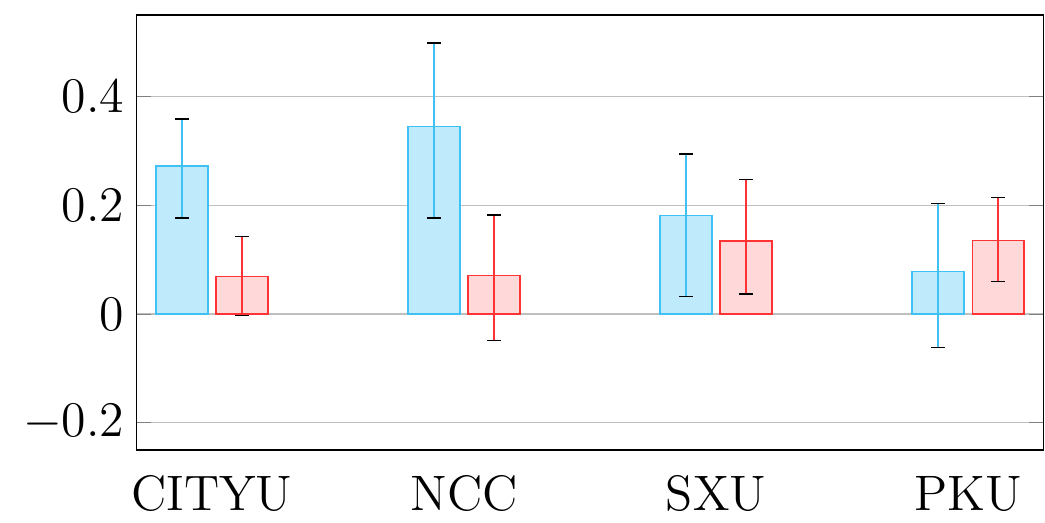} 
    }  \hspace{-0.15em}  
    \subfloat[NER]{
    \includegraphics[width=0.35\linewidth]{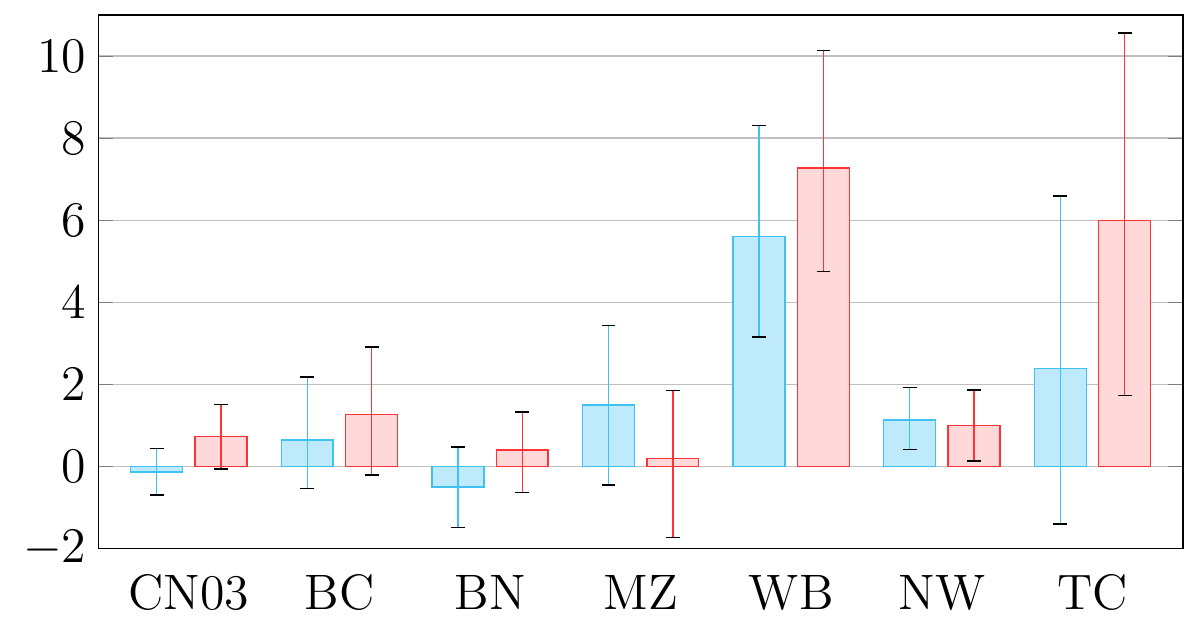} 
    }   \hspace{-0.15em}  
    \subfloat[Chunk \& POS]{
    \includegraphics[width=0.153\linewidth]{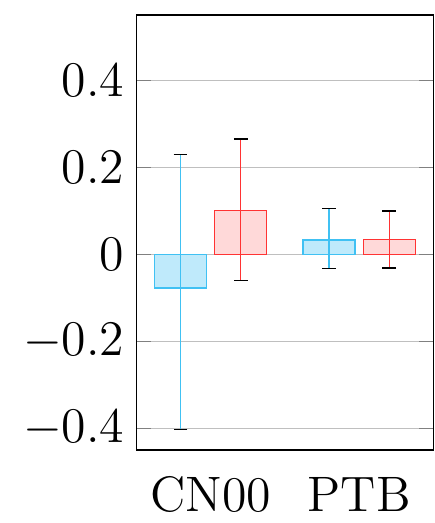} 
    }  
    \caption{Illustration of the relative improvement ($\%$) achieved by two larger-context methods (i.e., \textit{seq} and \textit{cPre-seq}) on four different tagging tasks. 
    The red and blue bars represent the improvements from \textit{seq} and \textit{cPre-seq}, respectively. 
    The error bars represent $95\%$ confidence intervals of the relative improvement that are computed based on Bootstrap method \cite{efron1986bootstrap}.
    } 
    \label{fig:larger_context_line}
\end{figure*}

\subsection{Exp-II: Effect of BERT}
\label{sec:bert-effect}
To answer the research question \textbf{Q2} ( \textit{Can the larger-context approach easily play to its strengths with the help of recently arising contextualized pre-trained models?}), we elaborate on how \textit{cPre-seq} and \textit{seq} aggregators influence the performance.

\paragraph{Results}
Fig.~\ref{fig:larger_context_line} illustrates the relative improvement achieved by two larger-context methods: \textit{seq} (blue bar) and \textit{cPre-seq} (red bar) on four different tagging tasks. 
We observe that:

\noindent 1) In general, aggregators equipped with BERT can not guarantee a better improvement, which is dataset-dependent.
2) Task-wisely, \textit{cPre-seq} can improve performance on all datasets on NER, Chunk, and POS tasks. By contrast, \textit{seq} is beneficial to all datasets on CWS task.
It could be attributed to the difference in language and characteristics of the task.
Specifically, for most non-CWS task datasets, \textit{cPre-seq} (7 out of 9 datasets) performs better than \textit{seq} ($p < 0.05$).

\section{Experiment: Why Does It Work?}
Experiments in this section are designed for the research questions \textbf{Q3}, interpreting where the gains of a larger-context approach come and why different datasets exhibit diverse improvements. 
To achieve this goal, we use the concept of interpretable evaluation \cite{fu2020interpretable} that allows us perform fine-grained evaluation of one or multiple systems.

\subsection{Attribute Definition}
\label{sec:attribute-definition}
The first step of interpretable evaluation is attribute definition.
The high-level idea is, given one attribute, the test set of each tagging task will be partitioned into several interpretable buckets based on it. And $F$1 score (accuracy for POS) will be calculated bucket-wisely. 
Next, we will explicate the general attributes we defined in this paper.

We first detail some notations to facilitate definitions of our attributes.
\uline{We define $x$ as a token and a bold form $\mathbf{x}$ as a span, which occurs in a test sentence $X = \mathrm{sent}(\mathbf{x})$.}
We additionally define two functions $\mathrm{oov(\cdot)}$ that counts the number out of training set words, and $\mathrm{ent(\cdot)}$ that tallies the number of entity words. 
Based on these notations, we introduce some feature functions that can compute different attributes for each span or token. Following, we will give the attribute definition of the NER.

\noindent
\textbf{Training set-independent Attributes}
\vspace{-5pt}
\begin{itemize}
    \item $\phi_{\texttt{eLen}}(\mathbf{x}) = |\mathbf{x}|$: \textit{entity span length}
    \vspace{-4pt}
    \item $\phi_{\texttt{sLen}}(\mathbf{x}) = |\mathrm{sent}(\mathbf{x})|$: \textit{sentence length} 
    \vspace{-4pt}
    \item $\phi_{\texttt{eDen}}(\mathbf{x}) = |\mathrm{ent}(sent(\mathbf{x}))|/\phi_{\texttt{sLen}}(\mathbf{x})$: \textit{entity density}
    \vspace{-4pt}
    \item $\phi_{\texttt{dOov}}(\mathbf{x}) = |\mathrm{oov}(\mathrm{sent}(\mathbf{x}))|/\phi_{\texttt{sLen}}(\mathbf{x})$: \textit{OOV density}    
\end{itemize}

\noindent
\textbf{Training set-dependent Attributes}
\vspace{-5pt}
\begin{itemize}
     \item $\phi_{\texttt{eFre}}(\mathbf{x}) = \mathrm{Fre}(\mathbf{x})$: \textit{entity frequency}
     \vspace{-4pt}
     \item $\phi_{\texttt{eCon}}(\mathbf{x}) = \mathrm{Con}(\mathbf{x})$: \textit{label consistency of entity}     
\end{itemize}

\noindent
where $\mathrm{Fre}(\mathbf{x})$ calculates the frequency of input $\mathbf{x}$ in the training set.
$\mathrm{Con}(\mathbf{x})$ quantify how consistently a given span is labeled with a particular label, and
$\mathrm{Con}(\mathbf{x})$ can be formulated as:

\begin{align}
 \mathrm{Con}(\mathbf{x}) &= \frac{|\{\varepsilon| \mathrm{lab}(\varepsilon)(\mathbf{x}), \forall \varepsilon \in \mathcal{E}^{tr}\}|}{|\mathrm{str}(\varepsilon) = \mathrm{str}(\mathbf{x}), \forall \varepsilon \in \mathcal{E}^{tr}\}|}, \\
    \label{eq:econ}
\mathrm{lab}(\varepsilon) &= \mathrm{lab}(\mathbf{x}) \cap  \mathrm{str}(\varepsilon) = \mathrm{str}(\mathbf{x}),
\end{align}

\noindent
where $\mathcal{E}^{tr}$ denotes entities in the training set, 
$\mathrm{lab}(\cdot)$ denotes the label of input span while $\mathrm{str}(\cdot)$ represents the surface string of input span. Similarly, we can extend the above two attributes to token-level, therefore obtaining $\phi_{\texttt{tFre}}(x)$ and $\phi_{\texttt{tCon}}(x)$.

Attributes for CWS task can be defined in a similar way.
Specifically, the entity (or token) in NER task corresponds to the word (or character) in CWS task. Note that we omit word density for CWS task since it equals to one for any sentence.

\subsection{Attribute Buckets}

We breakdown all test examples into different attribute buckets according to the given attribute. 
Take entity length (\texttt{eLen}) attribute of NER task as an example, first, we calculate each test sample's entity length attribute value. Then, divide the test entities into $N$ attribute buckets ($N=4$ by default) where the numbers of the test samples in all attribute intervals (buckets) are equal, and calculate the performance for those entities falling into the same bucket.

\subsection{Exp-I: Breakdown over Attributes}
\label{sec:exp-I}
\begin{figure*}
    \centering 
     \subfloat[{eCon}]{
    \includegraphics[width=0.133\linewidth]{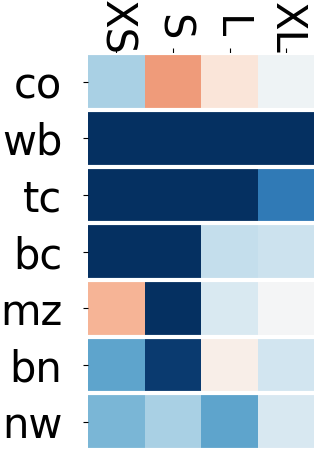}
    } 
    \subfloat[{tCon}]{
    \includegraphics[width=0.095\linewidth]{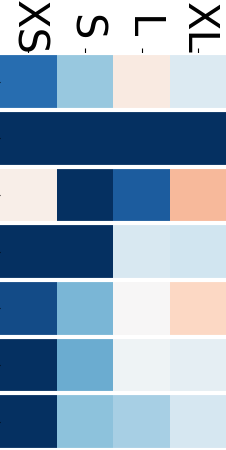}
    } 
    \subfloat[{eFre}]{
    \includegraphics[width=0.095\linewidth]{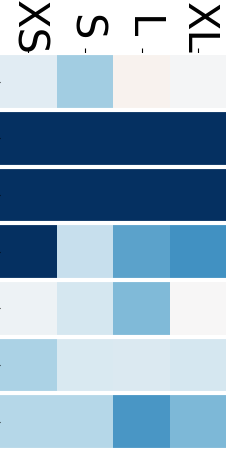}
    }  
    \subfloat[{tFre}]{
    \includegraphics[width=0.095\linewidth]{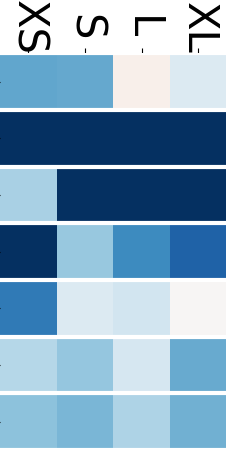} 
    } 
    \subfloat[{eDen}]{
    \includegraphics[width=0.095\linewidth]{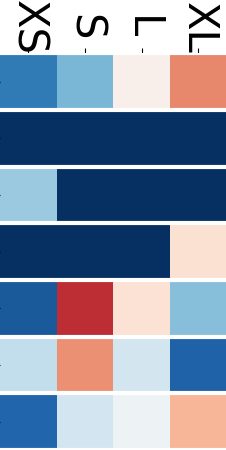}
    } 
    \subfloat[{eLen}]{
    \includegraphics[width=0.095\linewidth]{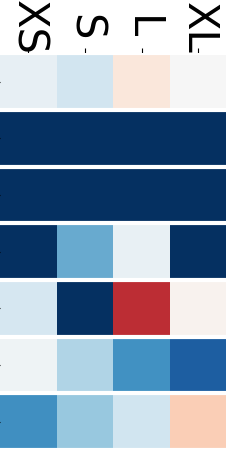}
    } 
    \subfloat[{sLen}]{
    \includegraphics[width=0.095\linewidth]{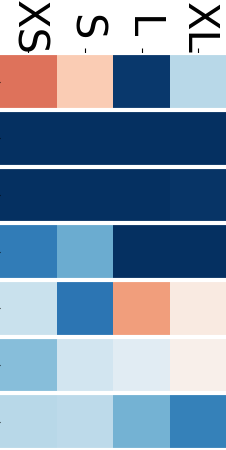}
    }
    \subfloat[{dOov}]{
    \includegraphics[width=0.166\linewidth]{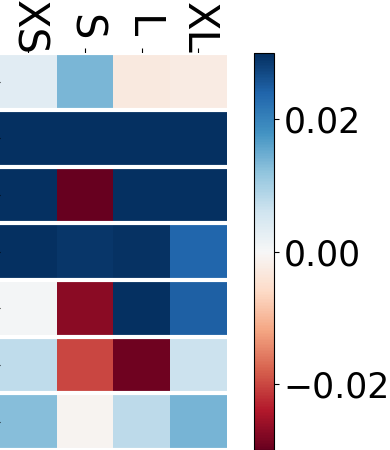}
    }  \\ 
    \subfloat[{wCon}]{
    \includegraphics[width=0.154\linewidth]{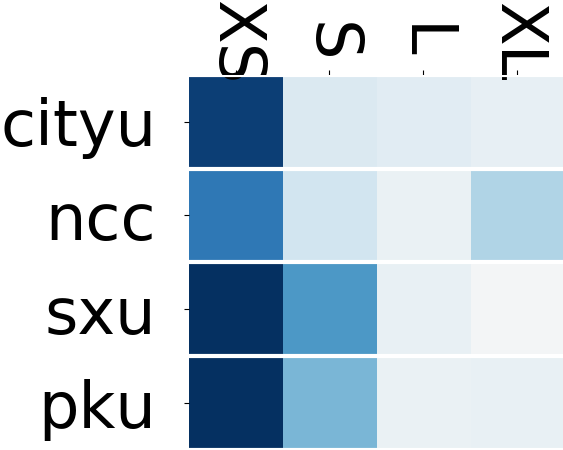}
    }  
    \subfloat[{cCon}]{ 
    \includegraphics[width=0.105\linewidth]{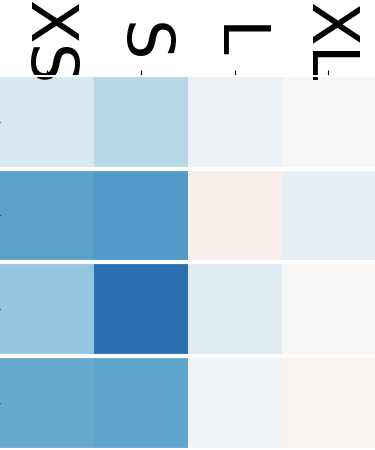}
    } 
    \subfloat[{wFre}]{
    \includegraphics[width=0.105\linewidth]{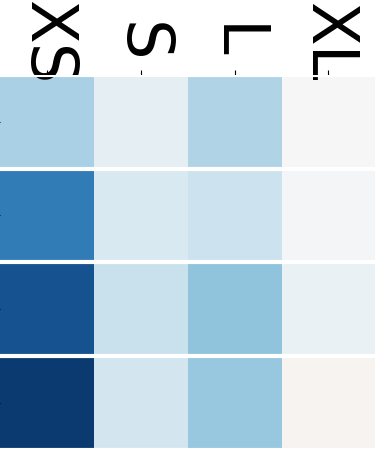}
    }  
    \subfloat[{cFre}]{
    \includegraphics[width=0.105\linewidth]{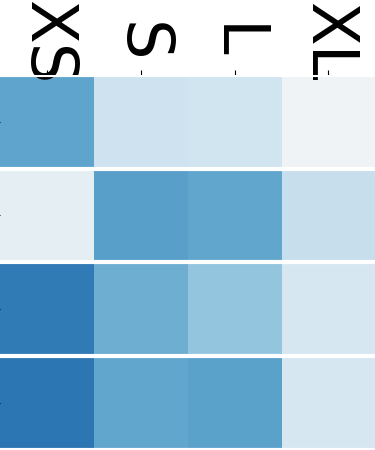} 
    } 
    \subfloat[{wLen}]{
    \includegraphics[width=0.115\linewidth]{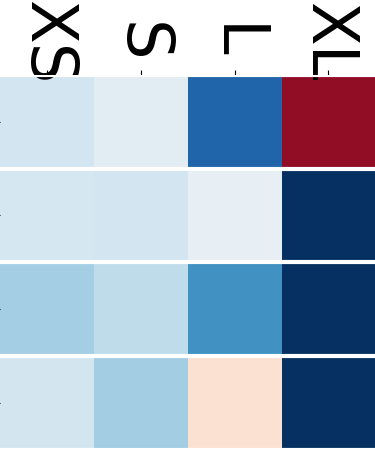}
    } 
    \subfloat[{sLen}]{
    \includegraphics[width=0.105\linewidth]{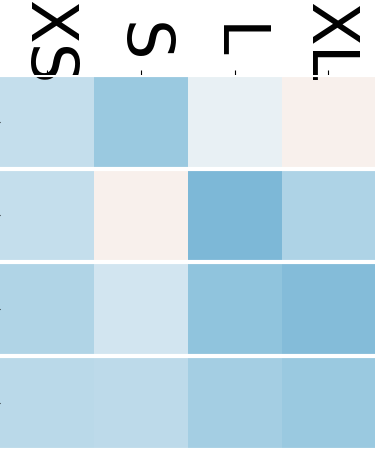}
    }
    \subfloat[{dOov}]{
    \includegraphics[width=0.175\linewidth]{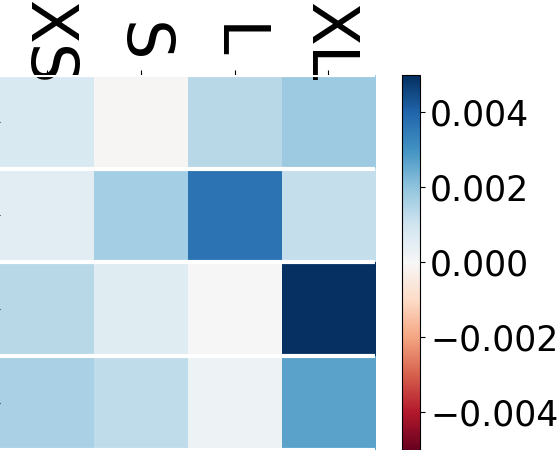}
    }  
    \caption{The relative increase ($\in [0,1]$) of the \textit{cPre-seq} larger-context training on NER (a$-$h) and CWS (i$-$o) tasks based on their evaluation attributes. ``\texttt{co}'' denotes the \texttt{CoNLL-2003} dataset. In order to facilitate observation, we divide the attribute value range into four categories: extra-small (XS), small (S), large (L), and extra-large (XL). The darker blue implies more significant improvement while the darker red suggests  larger-context leads to worse performance. 
    For the attribute name, ``\texttt{e}'', ``\texttt{t}'', ``\texttt{w}'', and ``\texttt{c}'' refers to ``\texttt{entity}'', ``\texttt{token}'', ``\texttt{word}'', and ``\texttt{character}'', respectively.
    } 
    \label{fig:larger_context_heat}
\end{figure*}

To investigate where the gains of the larger-context training come, we conduct a fine-grained evaluation with the evaluation attributes defined in Sec.~\ref{sec:attribute-definition}. We use the \textit{cPre-seq} larger-context aggregation method as the base model. Fig.~\ref{fig:larger_context_heat} shows the relative improvement of the \textit{cPre-seq} larger-context aggregation method in NER ($7$ datasets) and CWS tasks ($4$ datasets). The relative improvement is the performance of  \textit{cPre-seq} larger-context tagging minus sentence-level tagging.

\paragraph{Results}
Our findings from Fig.~\ref{fig:larger_context_heat} are:

\noindent 1) Test spans with lower label consistency can benefit much more from the larger-context training. As shown in  Fig.~\ref{fig:larger_context_heat} (a,b,i,j), test spans with lower label consistency (NER:\texttt{eCon,tCon=S/XS}, CWS: \texttt{wCon,cCon=S/XS}) can achieve higher relative improvement using the larger-context training, which holds for both NER and CWS tasks. 

\noindent 2)  NER task has achieved more gains on lower and higher-frequency test spans, while CWS task obtains more gains on lower-frequency test spans.
As shown in Fig.~\ref{fig:larger_context_line} (c,d,k,l), in NER task, test spans with higher or lower frequency (NER:\texttt{eFre=XS/XL;tFre=XS/XL}) will achieve larger improvements with the help of more contextual sentences; while for the CWS task, only the test spans with lower frequency will achieve more gains.

\noindent 3) Test spans of NER task with lower entity density have obtained larger improvement with the help of a larger-context training.
In terms of entity density shown in Fig.~\ref{fig:larger_context_heat} (e), an evaluation attribute specific to the NER task, the larger-context training is not good at dealing with the test spans with high entity density (NER:\texttt{eDen=XL/L}), while doing well in test spans with low entity density (NER:\texttt{eDen=XS/S}). 

\noindent 4) Larger-context training can achieve more gains on short entities in NER task while long words in CWS task.
As shown in 
Fig.~\ref{fig:larger_context_heat} (f,m), 
the dark blue boxes can be seen in the short entities (\texttt{eLen=XS/S}) of NER task, and long words (\texttt{wLen=XL/L}) of CWS task.

\noindent 5) Both NER and CWS tasks will achieve more gains on spans with higher OOV density.
For the OOV density shown in Fig.~\ref{fig:larger_context_line} (h,o), the test spans with higher OOV density (NER,CWS:\texttt{dOov=L/XL}) will achieve more gains from the larger-context training, which holds for both NER and CWS tasks.

\renewcommand\tabcolsep{1.2pt}
\begin{table*}[htb]
  \centering \footnotesize
    \begin{tabular}{p{0.4cm}lcccccccc}
    \toprule
         & \textbf{Attr.} & \texttt{eCon} & \texttt{tCon} & \texttt{eFre} & \texttt{tFre} & \texttt{eLen} & \texttt{dOov} & \texttt{sLen} & \texttt{eDen} \\
    \midrule
    \multicolumn{2}{c}{\multirow{4}[2]{*}{\textbf{$ \zeta_p$}}} & \multirow{4}[2]{*}{\includegraphics[scale=0.13]{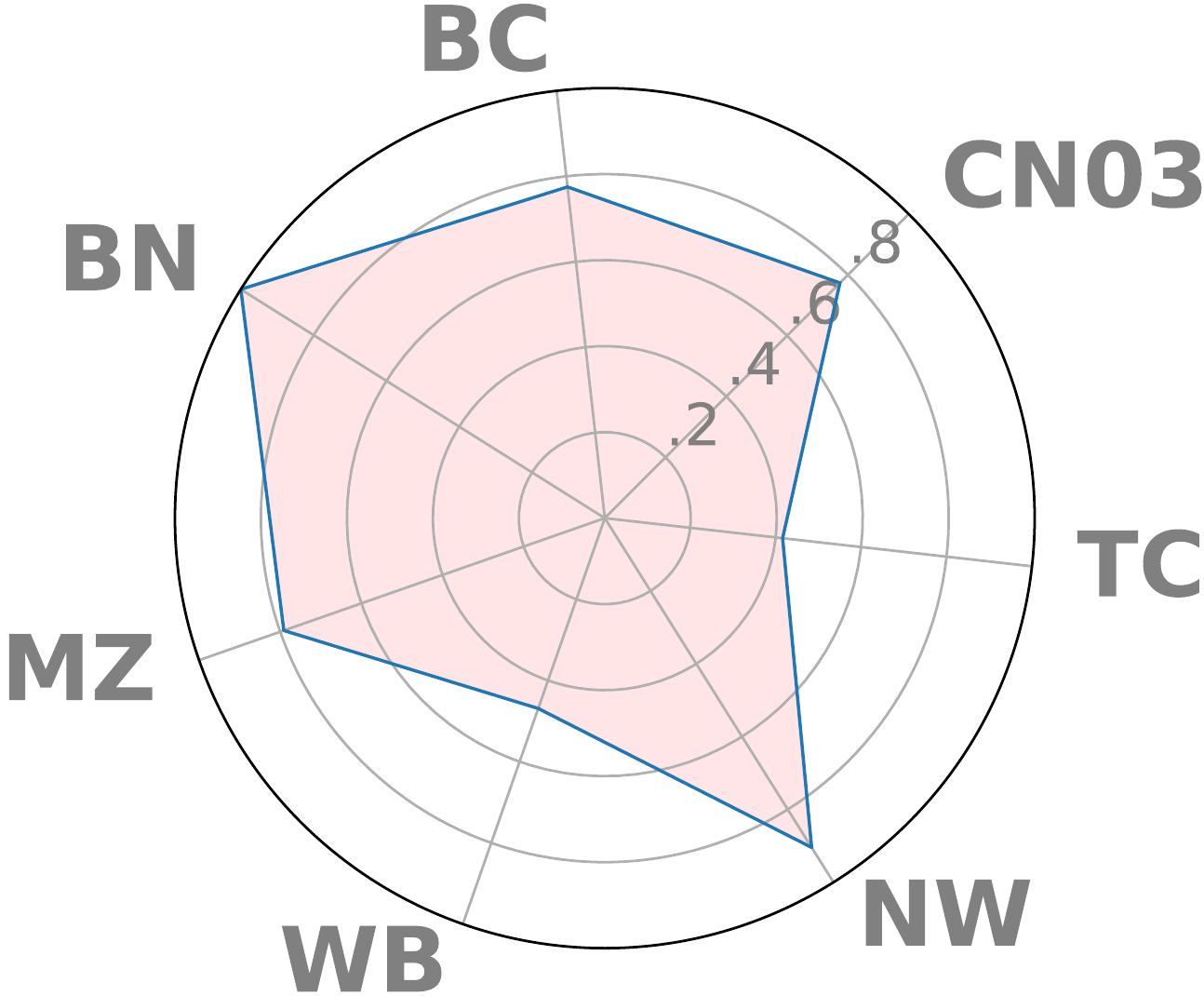}} & 
    \multirow{4}[2]{*}{\includegraphics[scale=0.13]{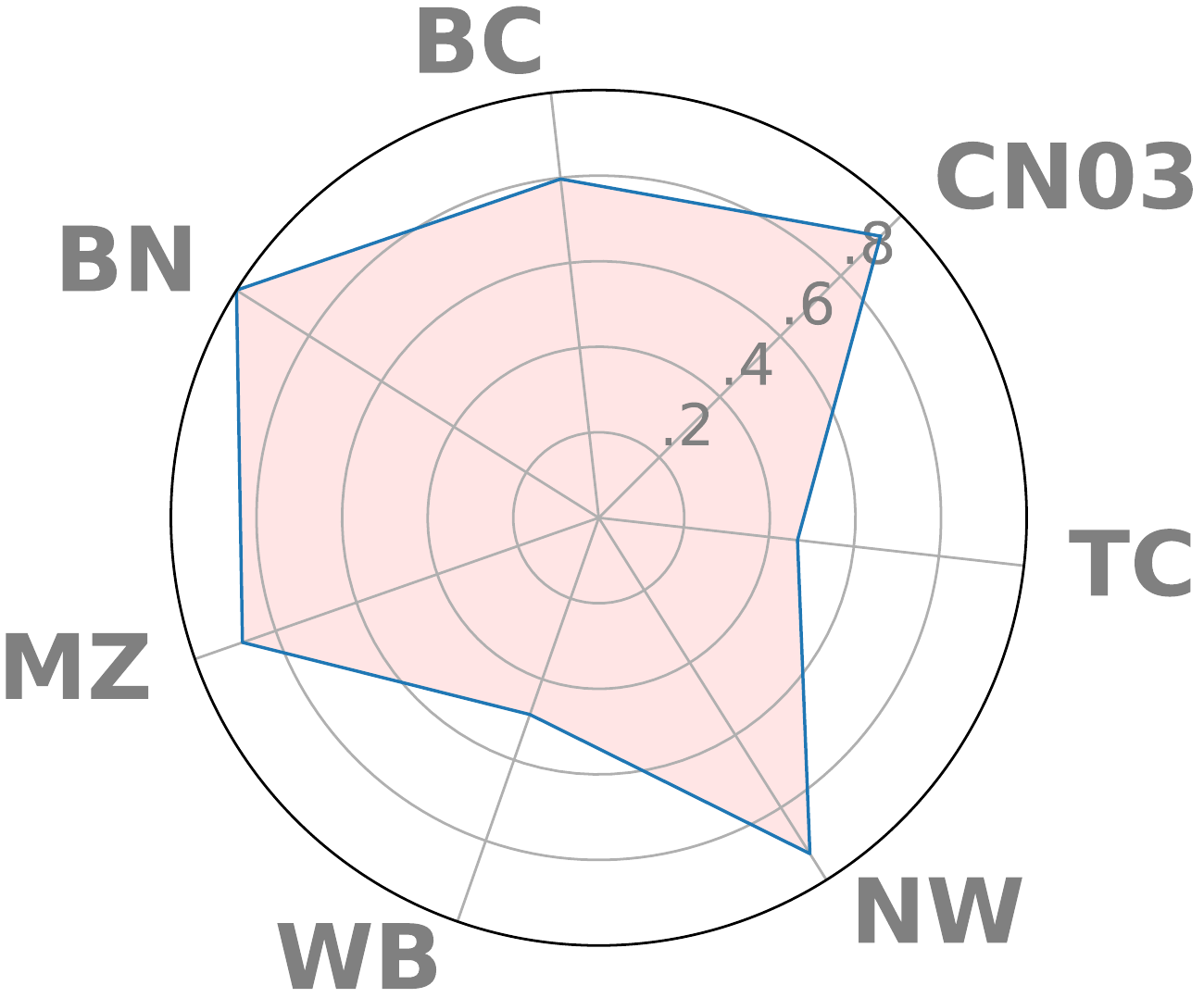}} & 
    \multirow{4}[2]{*}{\includegraphics[scale=0.13]{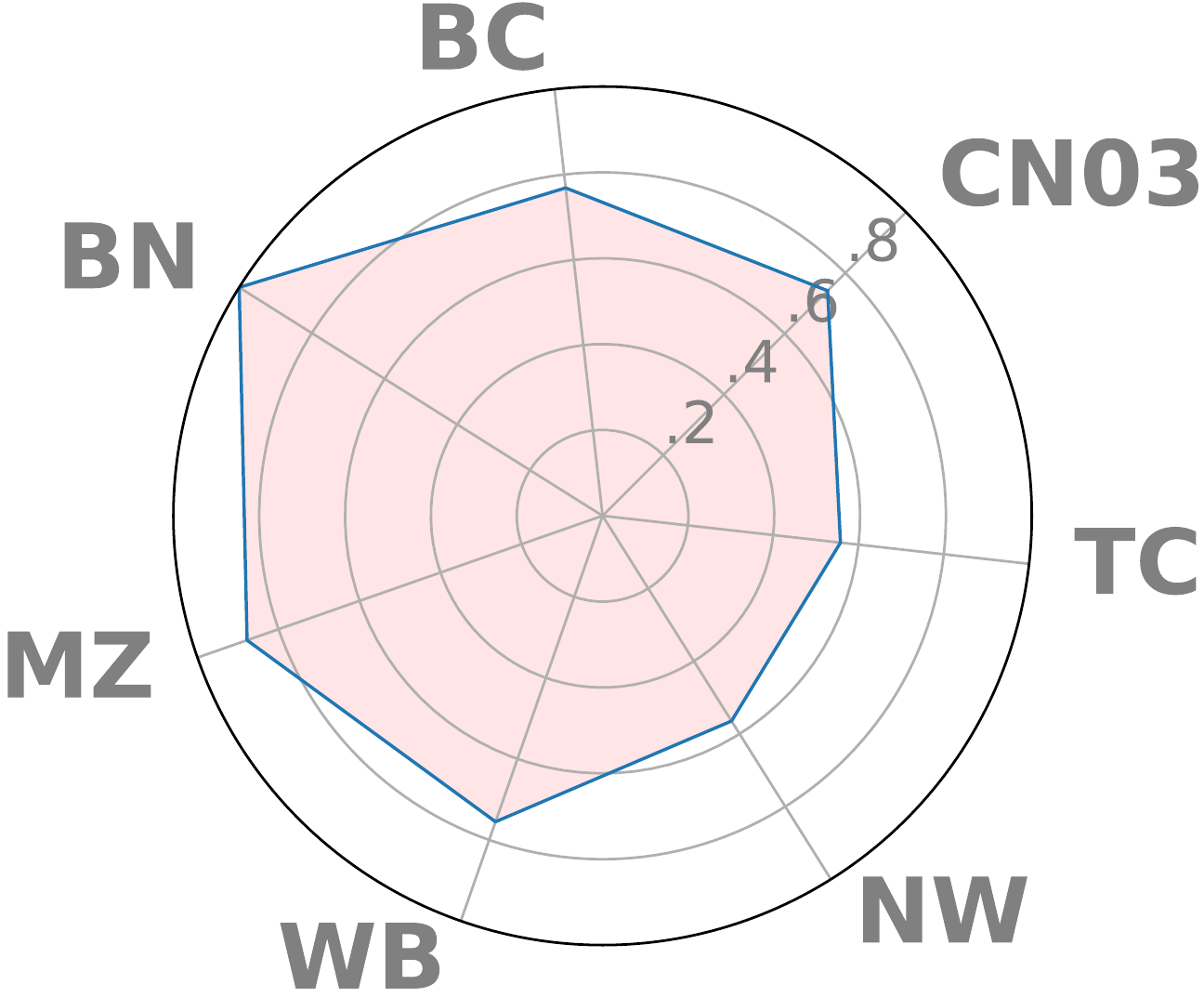}} & 
    \multirow{4}[2]{*}{\includegraphics[scale=0.13]{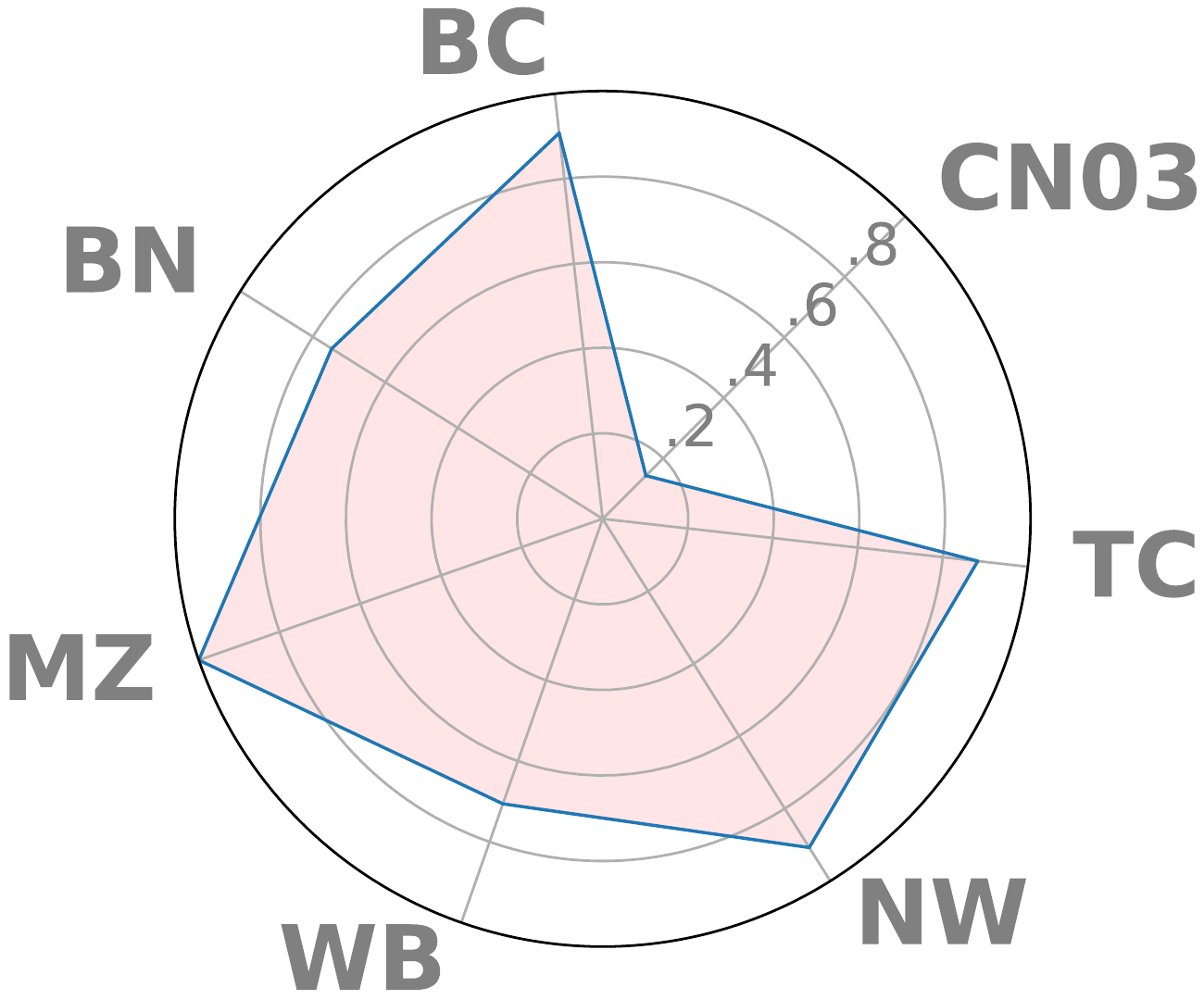}} & 
    \multirow{4}[2]{*}{\includegraphics[scale=0.13]{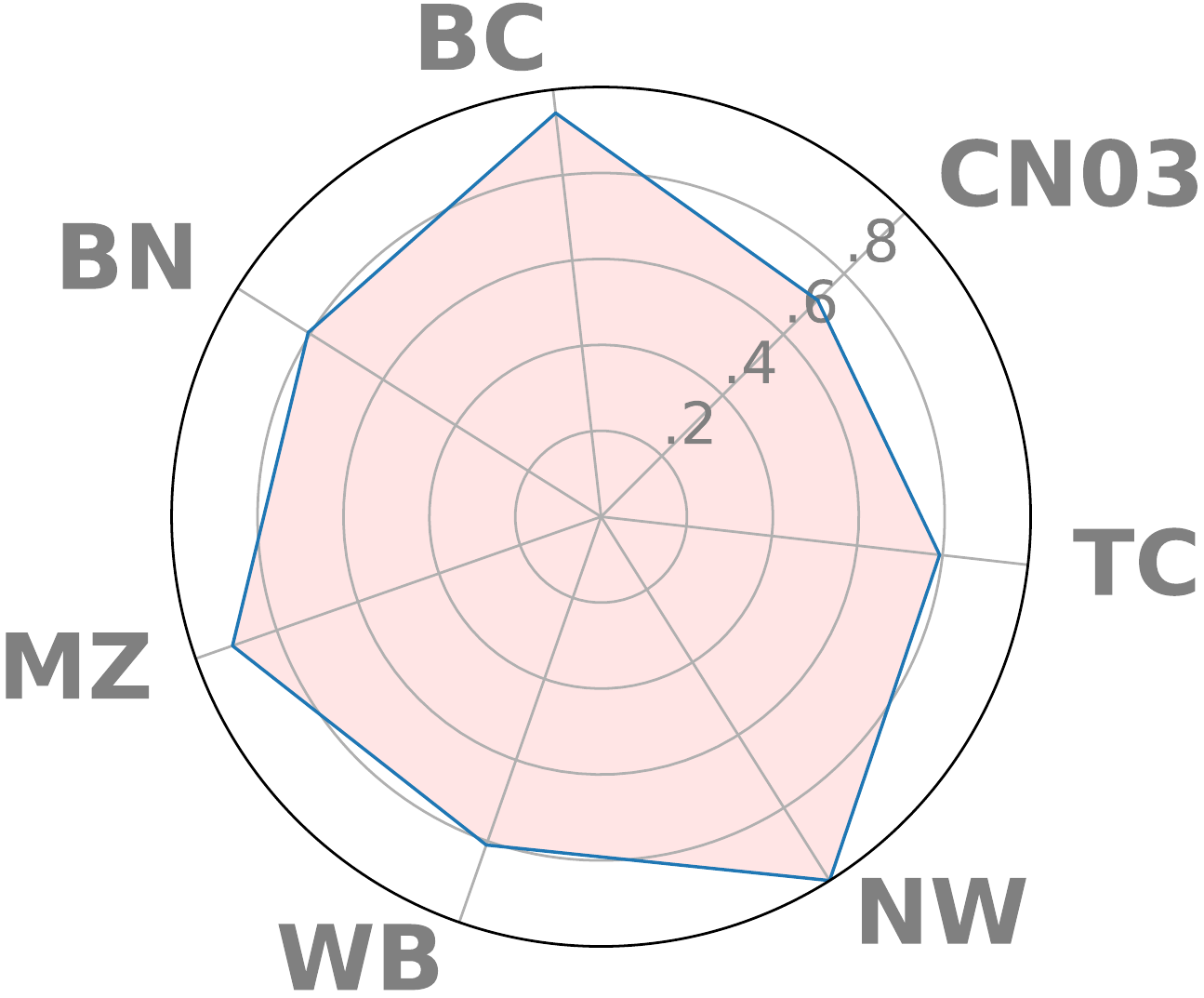}} & 
    \multirow{4}[2]{*}{\includegraphics[scale=0.13]{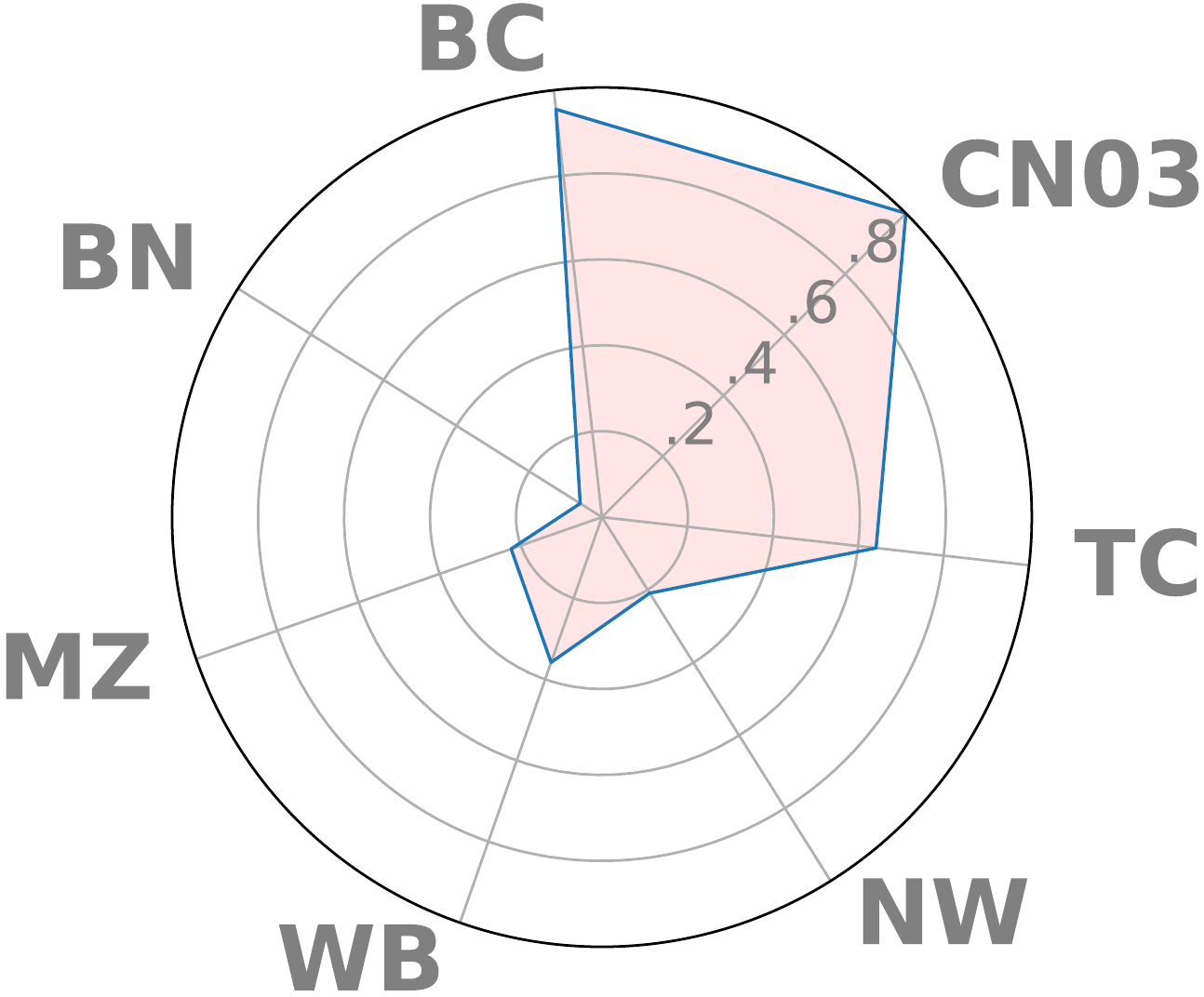}} & 
    \multirow{4}[2]{*}{\includegraphics[scale=0.13]{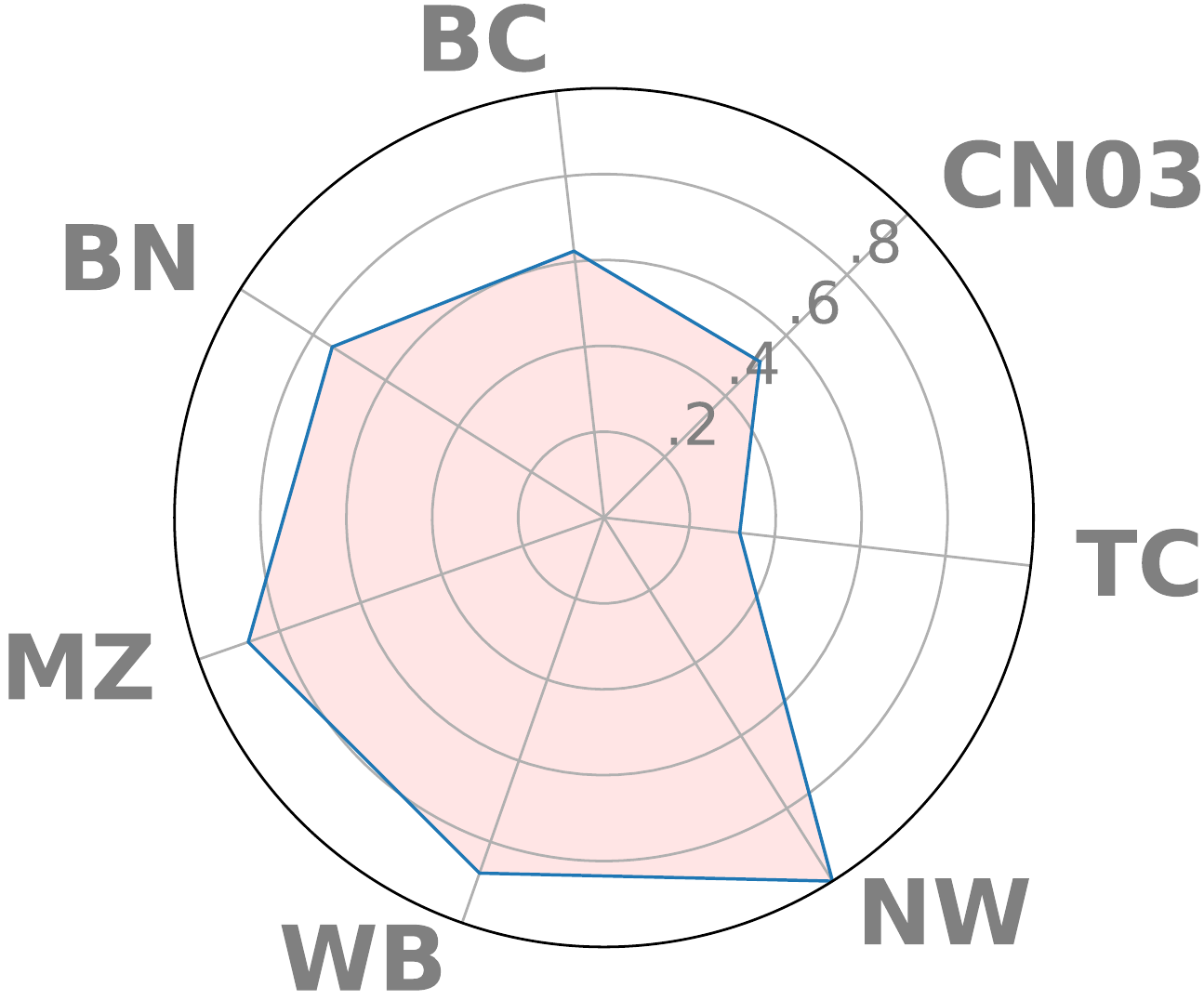}} & 
    \multirow{4}[2]{*}{\includegraphics[scale=0.13]{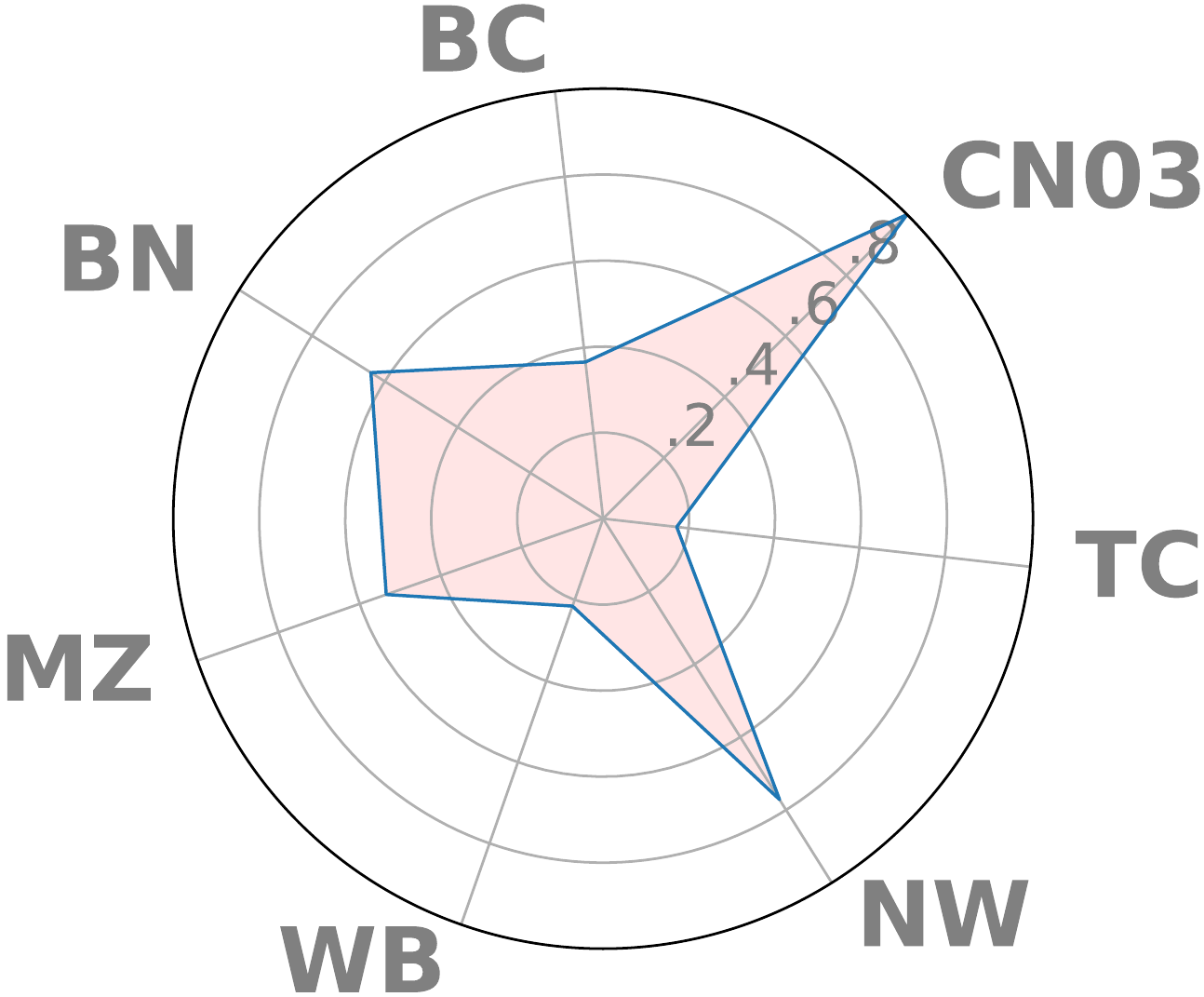}} \\
         \\  \\ \\ 
    \midrule
    \multirow{4}[2]{*}{\textbf{$\rho$}} & \textit{bow} & \textcolor{cadetgrey}{-0.179}  & -0.607  & \textcolor{cadetgrey}{0.143}  & 0.396  & 0.643  & \textcolor{cadetgrey}{-0.036}  & \textcolor[rgb]{0.0, 0.5, 0.0}{0.468}  & -0.571  \\
    & \textit{graph} & -0.571  & \textcolor{cadetgrey}{-0.143}  & -0.393  & \textcolor[rgb]{0.0, 0.5, 0.0}{-0.919}  & -0.643  & 0.286  & -0.288  & \textcolor{cadetgrey}{-0.107}  \\
    & \textit{seq} & -0.643  & \textcolor[rgb]{0.0, 0.5, 0.0}{-0.857}  & -0.429  & \textcolor{cadetgrey}{0.162}  & \textcolor{cadetgrey}{0.143}  & \textcolor{cadetgrey}{0.071}  & \textcolor{cadetgrey}{0.180}  & \textcolor[rgb]{0.0, 0.5, 0.0}{-0.714}  \\
    & \textit{cPre} & \textcolor[rgb]{0.0, 0.5, 0.0}{-0.714}  & -0.750  & \textcolor[rgb]{0.0, 0.5, 0.0}{-0.571}  & -0.306  & \textcolor{cadetgrey}{0.000}  & \textcolor[rgb]{0.0, 0.5, 0.0}{0.357}  & \textcolor{cadetgrey}{-0.180}  & -0.643  \\
    \bottomrule
    \end{tabular}%
 \caption{Illustration of measures $\zeta_p$ in seven datasets (\texttt{CN03}, \texttt{TC}, \texttt{NW}, \texttt{WB}, \texttt{MZ}, \texttt{BN}, \texttt{BC}) with respect to eight attributes (e.g., \texttt{eCon}) and correlation measure $\rho$ in NER task. 
 A higher absolute value $\rho$ (e.g \texttt{|-0.714|})  represents the improvement of the corresponding aggregator (e.g., \textit{seq}) heavily correlates with corresponding attribute (e.g. \texttt{eCon}). 
 The number with the highest absolute value of each column is colored by green.
 ``cPre'' represents ``cPre-seq'' and the values in grey denote correlation values do not pass a significance test ($p=0.05$).
 ``Attr.'' denotes attributes.
 }
    \label{tab:spearman-data-metric}%
\end{table*}%

\subsection{Exp-II: Quantifying and Understanding Dataset Bias}
\label{sec:exp-II}

Different datasets (e.g. \texttt{CN03}) may match different information aggregators (e.g. \textit{cPre-seq}).
Figuring out how different datasets influence the choices of aggregators is a challenging task. We try to approach this goal by (i) designing diverse measures that can characterize a given dataset from different perspectives, (ii) analyzing the correlation between different dataset properties and improvements brought by different aggregators.
\paragraph{Dataset-level Measure}
Given a dataset $\mathcal{E}$ and an attribute $p$ as defined in Sec.~\ref{sec:attribute-definition}, the data-level measure can be defined as:

\begin{equation}
    \zeta_p(\mathcal{E}) = \frac{1}{|\mathcal{E}^{te}|} \sum_{\varepsilon \in \mathcal{E}^{te}}\phi_p(\varepsilon) 
    \label{eq:zeta},
\end{equation}

\noindent
where $\mathcal{E}^{te} \in \mathcal{E}$ is a test set that contains entities/tokens in the NER task or word/character in the CWS task. $\phi_{p}(\cdot)$ is a function (as defined in Sec.~\ref{sec:attribute-definition}) that computes the attribute value for a given span.
For example, $\zeta_{\texttt{sLen}}(\texttt{CN03})$ represents the average sentence length of \texttt{CN03}'s test set.

\paragraph{Correlation Measure}
Statistically, we define a variable of $\rho$ to quantify the correlation between a dataset-level attribute and the relative improvement of an aggregator: $    \rho = \mathrm{Spearman}(\zeta_p, f_y)$,
where $\mathrm{Spearman}$ denotes the Spearman's rank correlation coefficient \cite{mukaka2012statistics}.
$\zeta_p$ represents dataset-level attribute values on all datasets with respect to attribute $p$ (e.g., \texttt{eLen}) while $f_y$ denotes the relative improvements of larger-context training on corresponding datasets with respect to a given aggregator $y$ (e.g., \textit{cPre-seq}).

\paragraph{Results}
\label{sec:choice}

Tab.~\ref{tab:spearman-data-metric} 
displays (using spider charts) measure $\zeta_p$~\footnote{The specific value  of $\zeta_p$ in NER and CWS task can be found in the appendix.}  of seven datasets with respect to diverse attributes, and correlation measure $\rho$ in the NER task.~\footnote{Analysis of other tasks can be found in our appendix section.}
Based on these correlations, which passed significantly test ($p<0.05$), between dataset-level measure (w.r.t a certain attribute, e.g. \texttt{eCon}) and gains from larger-context training (w.r.t an aggregator, e.g. \textit{seq}), we can obtain that:

\noindent (1) Regarding the \textit{cPre-seq} aggregator, it negatively correlated with $\zeta_{\texttt{eCon}}$, $\zeta_{\texttt{tCon}}$, $\zeta_{\texttt{eFre}}$, and $\zeta_{\texttt{eDen}}$ with larger correlation values.  Therefore, the \textit{cPre-seq} aggregator is more appropriate to deal with \texttt{WB}, \texttt{TC}, \texttt{BC} and \texttt{NW} datasets,
since these four datasets have a lower value of $\zeta_p$ with respect to the attribute \texttt{eCon} (\texttt{TC,WB}), \texttt{tCon} (\texttt{TC}, \texttt{WB}), \texttt{eFre} (\texttt{NW}, \texttt{TC}), and \texttt{eDen} (\texttt{BC}, \texttt{WB}, \texttt{TC}). 
Additionally, since the \textit{cPre-seq} aggregator obtains the highest positive correlation with $\zeta_{\texttt{dOov}}$, and $\zeta_{\texttt{dOov}}(\texttt{CN03})$, as well as $\zeta_{\texttt{dOov}}(\texttt{BC})$, achieve the highest value, \textit{cPre-seq} aggregator is suitable for \texttt{CN03} and \texttt{BC}.

\noindent (2) Regarding the \textit{seq} aggregator, 
it negatively correlated with $\zeta_{\texttt{eCon}}$, $\zeta_{\texttt{tCon}}$, and $\zeta_{\texttt{eDen}}$. Therefore, the \textit{seq} aggregator is better at dealing with datasets \texttt{WB}, \texttt{TC}, and \texttt{BC}, since these datasets are with lower $\zeta_p$ value on one of the attributes (\texttt{eCon}, \texttt{tCon}, and \texttt{eDen}).

\paragraph{Takeaways:}
We can conduct a similar analysis for \textit{bow} and \textit{graph} aggregators. Due to limited pages, we detail them in our appendix and highlight the suitable NER datasets for each aggregator as follows.

\vspace{-7pt}
\begin{itemize}
    \item [(1)] \textit{bow}: \texttt{WB}, \texttt{TC}, \texttt{NW}, \texttt{MZ}, \texttt{BC}.
    \vspace{-6pt}
    \item [(2)] \textit{graph}: \texttt{WB}, \texttt{TC}, \texttt{BN}, \texttt{CN03}.  
    \vspace{-6pt}
    \item [(3)] \textit{seq}: \texttt{WB}, \texttt{TC},  \texttt{BC}.
    \vspace{-6pt}
    \item [(4)] \textit{cPre-seq}: \texttt{CN03}, \texttt{WB}, \texttt{TC}, \texttt{BC}, \texttt{NW}.
\end{itemize}

\section{Adapting to Top-Scoring Systems}
\label{sec:top-systems}
Beyond the above quantitative and qualitative analysis of our instantiated typical tagging models (Sec.\ref{sec:taggmodels}), we are also curious about how well modern top-scoring tagging systems perform when equipped with larger-context training.

To this end, we choose the NER task as a case study and first re-implement existing top-performing models for different NER datasets separately, and then adapt larger-context approach to them based on the \textit{seq} or \textit{cPre-seq} aggregator,\footnote{Training all four aggregators for all tagging tasks is much more costly and here we choose these two since they can obtain better performance at a relatively lower cost.} which has shown superior performance in our above analysis. 

\paragraph{Settings} We collect five top-scoring tagging systems~\cite{luo2020hierarchical,lin2019reliability,chen2019grn,yan2019tener,akbik2018contextual} that are most recently proposed \footnote{We originally aimed to select more (10 systems) but suffer from reproducibility problems \cite{pineau2020improving}, even after contacting the first authors.}.
Among these five models,  regarding \newcite{akbik2018contextual}, we use \textit{cPre-seq} aggregator for the larger-context training, since this model originally relies on a contextualized pre-trained layer. Besides, from above analysis in Sec.~\ref{sec:choice} we know the suitable datasets for \textit{cPre-seq} aggregator: \texttt{CN03}, \texttt{WB}, \texttt{TC}, \texttt{BC}, and \texttt{NW}.
Regarding the other four models, we use the \textit{seq} aggregator for the larger-context training and the matched datasets are: \texttt{WB}, \texttt{TC}, and \texttt{BC}.

\paragraph{Results}
Tab.~\ref{tab:exist-model-result} shows the relative improvement of larger-context training on five modern top-scoring models in the NER task. We observe that the larger-context training has achieved consistent gains on all chosen datasets, which holds for both \textit{seq} and \textit{cPre-seq} aggregators. Notably, the larger-cotext training achieves sharp improvement on \texttt{WB}, which holds for all the five top-scoring models. For example, with the help of larger-context training, the performance can be improved significantly using  \newcite{akbik2018contextual}  and  \textbf{7.18} $F1$ score using \newcite{luo2020hierarchical}.  
This suggests that modern top-scoring NER systems can also benefit from larger-context training.

\renewcommand\tabcolsep{1.6pt}
\begin{table}[!htb]
  \centering  \scriptsize 
    \begin{tabular}{ccrrccccc}
    \toprule
    \multirow{2}[4]{*}{\textbf{Models}} & \multicolumn{3}{c}{\textbf{Aggregator}} & \multicolumn{5}{c}{\textbf{Datasets}} \\
\cmidrule{5-9}          & \textit{norm} & \multicolumn{1}{c}{\textit{seq}} & \multicolumn{1}{c}{\textit{cPre}} & \texttt{BC} & \texttt{WB} & \texttt{TC} & \texttt{CN03} & \texttt{NW} \\
    \midrule
    \multirow{2}[2]{*}{\newcite{luo2020hierarchical}} &  $\surd$     &       &       & 78.78  & 62.38  & 65.56  & -       & -  \\
          &       &  $\surd$     &       & \textcolor{green}{+0.54}  & \textcolor{green}{+7.18}  & \textcolor{green}{+1.7}   &       &  \\
    \midrule
    \multirow{2}[2]{*}{\newcite{lin2019reliability}} &  $\surd$     &       &       & 77.80  & 63.16  & 65.19  & -       & -  \\
          &       & $\surd$      &       & \textcolor{green}{+2.94}  & \textcolor{green}{+5.07}  & \textcolor{green}{+2.25}  &       &  \\
    \midrule
    \multirow{2}[2]{*}{\newcite{chen2019grn}} &  $\surd$     &       &       & 77.50  & 66.51  & 65.49  & -        & -  \\
          &       & $\surd$      &       & \textcolor{green}{+1.96}  & \textcolor{green}{+3.98}  & \textcolor{green}{+0.89}  & -       & -  \\
    \midrule
    \multirow{2}[2]{*}{\newcite{yan2019tener}} &  $\surd$     &       &       & 81.29  & 65.05  & 67.92  & 92.17  & 90.37  \\
          &       & $\surd$      &       & \textcolor{green}{+0.16}  & \textcolor{green}{+6.79}  & \textcolor{green}{+3.03}  & \textcolor{green}{+0.04}  & \textcolor{green}{+1.11} \\
    \midrule
    \multirow{2}[2]{*}{\newcite{akbik2018contextual} } &  $\surd$     &       &       & 81.13  & 64.79  & 69.00  & 93.03  & 90.76  \\
          &       &       & $\surd$      & \textcolor{green}{+1.12}  & \textcolor{green}{+10.78} & \textcolor{green}{+2.12}  & \textcolor{green}{+0.05}  & \textcolor{green}{+1.03} \\
    \bottomrule
    \end{tabular}%
    \vspace{-7pt}
    \caption{The relative improvement of larger-context training on top-scoring models in the NER task.~\footnotetext{Since \newcite{luo2020hierarchical} only released the code without BERT, and the results we re-implemented are quite lower than the result reported in their paper, we only report the model without using BERT.} ``cPre'' represents ``cPre-seq''. ``\textbf{norm}'' denotes the normal setting ($K = 1$).  The testing datasets are chosen based on the analysis in Sec.~\ref{sec:choice}. }
  \label{tab:exist-model-result}%
\end{table}%

\section{Related Work}
Our work touches the following research topics for tagging tasks.

\noindent
\textbf{Sentence-level Tagging}
Existing works have achieved impressive performance at sentence-level tagging by extensive structural explorations with different types of neural components.
Regarding sentence encoders, recurrent neural nets~\cite{huang2015bidirectional,chiu2015named,ma2016end,lample2016neural,li2019unified,lin2020triggerner} and convolutional neural nets~\cite{strubell2017fast,yang2018design,chen2019grn,fu2020interpretable} were widely used while transformer were also studied to get sentential representations~\cite{yan2019tener,yu2020improving}.
Some recent works consider the NER as a span classification~\cite{li2019unified,jiang2019generalizing,mengge2020coarse,ouchi2020instance} task, unlike most works that view it as a sequence labeling task.
To capture morphological information, some previous works introduced a character or subword-aware encoders with unsupervised pre-trained knowledge ~\cite{peters2018deep,akbik2018contextual,devlin2018bert,akbik2019pooled,yang2019xlnet,lan2019albert}.

\noindent
\textbf{Document-level Tagging}
Document-level tagging introduced more contextual features to improve the performance of tagging.
Some early works introduced non-local information~\cite{finkel2005incorporating,krishnan2006effective} to enhance traditional machine learning methods (e.g., CRF~\cite{lafferty2001conditional}) and achieved impressive results.
\newcite{qian2018graphie:,wadden2019entity} built graph representation based on the broad dependencies between words and sentences.
\newcite{luo2020hierarchical} proposed to use a memory network to record the document-aware information. 
Besides, document-level features was introduced by different domains to alleviate label inconsistency problems, such as news NER~\cite{hu2020leveraging,hu2019document}, chemical NER~\cite{luo2018attention}, disease NER~\cite{xu2019document}, and Chinese patent~\cite{li2014effective,li2016towards}.
Compared with these works, instead of proposing a novel model, we focus on investigating when and why the larger-context training, as a general strategy, can work.

\paragraph{Interpretability and Robustness of Sequence Labeling Systems}
Recently, there is a popular trend that aims to  (i) perform a glass-box analysis of sequence labeling systems \cite{fu2020rethinking,agarwal2020interpretability}, understanding their generalization ability and quantify robustness \cite{fu-etal-2020-rethinkcws}, (ii) interpretable evaluation of them \cite{fu2020interpretable}, making it possible to know what a system is good/bad at and where a system outperforms another, (iii) reliable analysis \cite{ye2021towards} for test set with fewer samples. 
Our work is based on the technique of interpretable evaluation, which provides a convenient way for us to diagnose different systems.

\section{Discussion}
\label{sec:discussion}
We summarize the main observations from our experiments and try to provide preliminary answers to our proposed research questions:

\noindent
(i) \textbf{\textit{How do different integration ways of larger-context information influence the system's performance?}} 
Overall, introducing larger-context information will bring gains regardless of the ways how to introduce it (e.g., \textit{seq}, \textit{graph}). Particularly, larger-context training with \textit{seq} aggregator can achieve better performance at lower training cost compared with \textit{graph} and \textit{bow} aggregators (Sec.~\ref{sec:structured-effect}).

\noindent
(ii) \textbf{\textit{Can the larger-context training easily play to its strengths with the help of contextualized pre-trained models?}} 
Yes for all datasets on NER, Chunk, and POS tasks.
By contrast, for CWS tasks, the aggregator without BERT (e.g., \textit{seq}) can achieve better improvement (Sec.~\ref{sec:bert-effect}).

\noindent
(iii) \textbf{\textit{Where does the gain of larger-context training come? And how do different characteristics of datasets affect the amount of gain?}} 
The source of gains, though, is dataset- and aggregator-dependent, a relatively consensus observation is that text spans with lower label consistency and higher OOV density can benefit a lot from larger-context training  (Sec.~\ref{sec:exp-I}).
Regarding different datasets, diverse aggregators are recommended in Sec.~\ref{sec:exp-II}.

\section*{Acknowledgments}
We would like to thank the anonymous reviewers for their valuable comments.
This work was supported by China National Key R$\&$D Program (No.2018YFC0831105).

\bibliography{naacl2021}

\begin{thebibliography}{50}
\expandafter\ifx\csname natexlab\endcsname\relax\def\natexlab#1{#1}\fi

\bibitem[{Agarwal et~al.(2020)Agarwal, Yang, Wallace, and
  Nenkova}]{agarwal2020interpretability}
Oshin Agarwal, Yinfei Yang, Byron~C Wallace, and Ani Nenkova. 2020.
\newblock Interpretability analysis for named entity recognition to understand
  system predictions and how they can improve.
\newblock \emph{arXiv preprint arXiv:2004.04564}.

\bibitem[{Akbik et~al.(2019)Akbik, Bergmann, and Vollgraf}]{akbik2019pooled}
Alan Akbik, Tanja Bergmann, and Roland Vollgraf. 2019.
\newblock Pooled contextualized embeddings for named entity recognition.
\newblock pages 724--728.

\bibitem[{Akbik et~al.(2018)Akbik, Blythe, and Vollgraf}]{akbik2018contextual}
Alan Akbik, Duncan Blythe, and Roland Vollgraf. 2018.
\newblock Contextual string embeddings for sequence labeling.
\newblock In \emph{Proceedings of the 27th International Conference on
  Computational Linguistics}, pages 1638--1649.

\bibitem[{Chen et~al.(2019)Chen, Lin, Ding, Lou, Zhang, and
  Karlsson}]{chen2019grn}
Hui Chen, Zijia Lin, Guiguang Ding, Jianguang Lou, Yusen Zhang, and Borje
  Karlsson. 2019.
\newblock Grn: Gated relation network to enhance convolutional neural network
  for named entity recognition.
\newblock In \emph{Proceedings of the AAAI Conference on Artificial
  Intelligence}, volume~33, pages 6236--6243.

\bibitem[{Chiu and Nichols(2015)}]{chiu2015named}
Jason P~C Chiu and Eric Nichols. 2015.
\newblock Named entity recognition with bidirectional lstm-cnns.
\newblock \emph{arXiv: Computation and Language}.

\bibitem[{Chiu and Nichols(2016)}]{chiu2016named}
Jason~PC Chiu and Eric Nichols. 2016.
\newblock Named entity recognition with bidirectional lstm-cnns.
\newblock \emph{Transactions of the Association for Computational Linguistics},
  4:357--370.

\bibitem[{Devlin et~al.(2018)Devlin, Chang, Lee, and
  Toutanova}]{devlin2018bert}
Jacob Devlin, Ming-Wei Chang, Kenton Lee, and Kristina Toutanova. 2018.
\newblock Bert: Pre-training of deep bidirectional transformers for language
  understanding.
\newblock \emph{arXiv preprint arXiv:1810.04805}.

\bibitem[{Durrett and Klein(2014)}]{durrett2014joint}
Greg Durrett and Dan Klein. 2014.
\newblock A joint model for entity analysis: Coreference, typing, and linking.
\newblock \emph{Transactions of the association for computational linguistics},
  2:477--490.

\bibitem[{Efron and Tibshirani(1986)}]{efron1986bootstrap}
Bradley Efron and Robert Tibshirani. 1986.
\newblock Bootstrap methods for standard errors, confidence intervals, and
  other measures of statistical accuracy.
\newblock \emph{Statistical science}, pages 54--75.

\bibitem[{Finkel et~al.(2005)Finkel, Grenager, and
  Manning}]{finkel2005incorporating}
Jenny~Rose Finkel, Trond Grenager, and Christopher Manning. 2005.
\newblock Incorporating non-local information into information extraction
  systems by gibbs sampling.
\newblock In \emph{Proceedings of the 43rd annual meeting on association for
  computational linguistics}, pages 363--370. Association for Computational
  Linguistics.

\bibitem[{Fu et~al.(2020{\natexlab{a}})Fu, Liu, and
  Neubig}]{fu2020interpretable}
Jinlan Fu, Pengfei Liu, and Graham Neubig. 2020{\natexlab{a}}.
\newblock \href {https://doi.org/10.18653/v1/2020.emnlp-main.489}
  {Interpretable multi-dataset evaluation for named entity recognition}.
\newblock In \emph{Proceedings of the 2020 Conference on Empirical Methods in
  Natural Language Processing (EMNLP)}, pages 6058--6069, Online. Association
  for Computational Linguistics.

\bibitem[{Fu et~al.(2020{\natexlab{b}})Fu, Liu, and Zhang}]{fu2020rethinking}
Jinlan Fu, Pengfei Liu, and Qi~Zhang. 2020{\natexlab{b}}.
\newblock Rethinking generalization of neural models: A named entity
  recognition case study.
\newblock In \emph{Proceedings of the AAAI Conference on Artificial
  Intelligence}, volume~34, pages 7732--7739.

\bibitem[{Fu et~al.(2020{\natexlab{c}})Fu, Liu, Zhang, and
  Huang}]{fu-etal-2020-rethinkcws}
Jinlan Fu, Pengfei Liu, Qi~Zhang, and Xuanjing Huang. 2020{\natexlab{c}}.
\newblock \href {https://doi.org/10.18653/v1/2020.emnlp-main.457}
  {{R}ethink{CWS}: Is {C}hinese word segmentation a solved task?}
\newblock In \emph{Proceedings of the 2020 Conference on Empirical Methods in
  Natural Language Processing (EMNLP)}, pages 5676--5686, Online. Association
  for Computational Linguistics.

\bibitem[{Ghaddar and Langlais(2018)}]{ghaddar2018robust}
Abbas Ghaddar and Philippe Langlais. 2018.
\newblock Robust lexical features for improved neural network named-entity
  recognition.
\newblock \emph{arXiv preprint arXiv:1806.03489}.

\bibitem[{Hu et~al.(2019)Hu, Dou, and Wen}]{hu2019document}
Anwen Hu, Zhicheng Dou, and Ji-rong Wen. 2019.
\newblock Document-level named entity recognition by incorporating global and
  neighbor features.
\newblock In \emph{China Conference on Information Retrieval}, pages 79--91.
  Springer.

\bibitem[{Hu et~al.(2020)Hu, Dou, Wen, and Nie}]{hu2020leveraging}
Anwen Hu, Zhicheng Dou, Jirong Wen, and Jianyun Nie. 2020.
\newblock Leveraging multi-token entities in document-level named entity
  recognition.

\bibitem[{Huang et~al.(2015)Huang, Xu, and Yu}]{huang2015bidirectional}
Zhiheng Huang, Wei Xu, and Kai Yu. 2015.
\newblock Bidirectional lstm-crf models for sequence tagging.
\newblock \emph{arXiv: Computation and Language}.

\bibitem[{Jiang et~al.(2019)Jiang, Xu, Araki, and
  Neubig}]{jiang2019generalizing}
Zhengbao Jiang, Wei Xu, Jun Araki, and Graham Neubig. 2019.
\newblock Generalizing natural language analysis through span-relation
  representations.
\newblock \emph{arXiv preprint arXiv:1911.03822}.

\bibitem[{Kipf and Welling(2016)}]{kipf2016semi}
Thomas~N Kipf and Max Welling. 2016.
\newblock Semi-supervised classification with graph convolutional networks.
\newblock \emph{arXiv preprint arXiv:1609.02907}.

\bibitem[{Krishnan and Manning(2006)}]{krishnan2006effective}
Vijay Krishnan and Christopher~D Manning. 2006.
\newblock An effective two-stage model for exploiting non-local dependencies in
  named entity recognition.
\newblock In \emph{Proceedings of the 21st International Conference on
  Computational Linguistics and the 44th annual meeting of the Association for
  Computational Linguistics}, pages 1121--1128. Association for Computational
  Linguistics.

\bibitem[{Lafferty et~al.(2001)Lafferty, McCallum, and
  Pereira}]{lafferty2001conditional}
John~D. Lafferty, Andrew McCallum, and Fernando C.~N. Pereira. 2001.
\newblock Conditional random fields: Probabilistic models for segmenting and
  labeling sequence data.
\newblock In \emph{Proceedings of the Eighteenth International Conference on
  Machine Learning}.

\bibitem[{Lample et~al.(2016)Lample, Ballesteros, Subramanian, Kawakami, and
  Dyer}]{lample2016neural}
Guillaume Lample, Miguel Ballesteros, Sandeep Subramanian, Kazuya Kawakami, and
  Chris Dyer. 2016.
\newblock Neural architectures for named entity recognition.
\newblock In \emph{Proceedings of NAACL-HLT}, pages 260--270.

\bibitem[{Lan et~al.(2019)Lan, Chen, Goodman, Gimpel, Sharma, and
  Soricut}]{lan2019albert}
Zhenzhong Lan, Mingda Chen, Sebastian Goodman, Kevin Gimpel, Piyush Sharma, and
  Radu Soricut. 2019.
\newblock Albert: A lite bert for self-supervised learning of language
  representations.
\newblock \emph{arXiv preprint arXiv:1909.11942}.

\bibitem[{Li and Xue(2014)}]{li2014effective}
Si~Li and Nianwen Xue. 2014.
\newblock Effective document-level features for chinese patent word
  segmentation.
\newblock 2:199--205.

\bibitem[{Li and Xue(2016)}]{li2016towards}
Si~Li and Nianwen Xue. 2016.
\newblock Towards accurate word segmentation for chinese patents.
\newblock \emph{arXiv: Computation and Language}.

\bibitem[{Li et~al.(2019)Li, Feng, Meng, Han, Wu, and Li}]{li2019unified}
Xiaoya Li, Jingrong Feng, Yuxian Meng, Qinghong Han, Fei Wu, and Jiwei Li.
  2019.
\newblock A unified mrc framework for named entity recognition.
\newblock \emph{arXiv preprint arXiv:1910.11476}.

\bibitem[{Lin et~al.(2020)Lin, Lee, Shen, Moreno, Huang, Shiralkar, and
  Ren}]{lin2020triggerner}
Bill~Yuchen Lin, Dong-Ho Lee, Ming Shen, Ryan Moreno, Xiao Huang, Prashant
  Shiralkar, and Xiang Ren. 2020.
\newblock Triggerner: Learning with entity triggers as explanations for named
  entity recognition.
\newblock \emph{arXiv preprint arXiv:2004.07493}.

\bibitem[{Lin et~al.(2019)Lin, Liu, Ji, Yu, and Han}]{lin2019reliability}
Ying Lin, Liyuan Liu, Heng Ji, Dong Yu, and Jiawei Han. 2019.
\newblock Reliability-aware dynamic feature composition for name tagging.
\newblock In \emph{Proceedings of the 57th Annual Meeting of the Association
  for Computational Linguistics}, pages 165--174.

\bibitem[{Luo et~al.(2018)Luo, Yang, Yang, Zhang, Wang, Lin, and
  Wang}]{luo2018attention}
Ling Luo, Zhihao Yang, Pei Yang, Yin Zhang, Lei Wang, Hongfei Lin, and Jian
  Wang. 2018.
\newblock An attention-based bilstm-crf approach to document-level chemical
  named entity recognition.
\newblock \emph{Bioinformatics}, 34(8):1381--1388.

\bibitem[{Luo et~al.(2020)Luo, Xiao, and Zhao}]{luo2020hierarchical}
Ying Luo, Fengshun Xiao, and Hai Zhao. 2020.
\newblock Hierarchical contextualized representation for named entity
  recognition.

\bibitem[{Ma and Hovy(2016)}]{ma2016end}
Xuezhe Ma and Eduard Hovy. 2016.
\newblock End-to-end sequence labeling via bi-directional lstm-cnns-crf.
\newblock In \emph{Proceedings of the 54th Annual Meeting of ACL}, volume~1,
  pages 1064--1074.

\bibitem[{Mengge et~al.(2020)Mengge, Yu, Zhang, Liu, Zhang, and
  Wang}]{mengge2020coarse}
Xue Mengge, Bowen Yu, Zhenyu Zhang, Tingwen Liu, Yue Zhang, and Bin Wang. 2020.
\newblock Coarse-to-fine pre-training for named entity recognition.
\newblock In \emph{Proceedings of the 2020 Conference on Empirical Methods in
  Natural Language Processing (EMNLP)}, pages 6345--6354.

\bibitem[{Mikolov et~al.(2013)Mikolov, Sutskever, Chen, Corrado, and
  Dean}]{mikolov2013distributed}
Tomas Mikolov, Ilya Sutskever, Kai Chen, Greg~S Corrado, and Jeff Dean. 2013.
\newblock Distributed representations of words and phrases and their
  compositionality.
\newblock In \emph{Advances in neural information processing systems}, pages
  3111--3119.

\bibitem[{Mukaka(2012)}]{mukaka2012statistics}
Mavuto Mukaka. 2012.
\newblock Statistics corner: A guide to appropriate use of correlation
  coefficient in medical research.
\newblock \emph{Malawi medical journal : the journal of Medical Association of
  Malawi}, 24(3):69--71.

\bibitem[{Ouchi et~al.(2020)Ouchi, Suzuki, Kobayashi, Yokoi, Kuribayashi,
  Konno, and Inui}]{ouchi2020instance}
Hiroki Ouchi, Jun Suzuki, Sosuke Kobayashi, Sho Yokoi, Tatsuki Kuribayashi,
  Ryuto Konno, and Kentaro Inui. 2020.
\newblock Instance-based learning of span representations: A case study through
  named entity recognition.
\newblock \emph{arXiv preprint arXiv:2004.14514}.

\bibitem[{Pennington et~al.(2014)Pennington, Socher, and
  Manning}]{pennington2014glove}
Jeffrey Pennington, Richard Socher, and Christopher~D Manning. 2014.
\newblock Glove: Global vectors for word representation.
\newblock \emph{Proceedings of the EMNLP}, 12:1532--1543.

\bibitem[{Peters et~al.(2018)Peters, Neumann, Iyyer, Gardner, Clark, Lee, and
  Zettlemoyer}]{peters2018deep}
Matthew Peters, Mark Neumann, Mohit Iyyer, Matt Gardner, Christopher Clark,
  Kenton Lee, and Luke Zettlemoyer. 2018.
\newblock Deep contextualized word representations.
\newblock In \emph{Proceedings of the 2018 Conference of the North American
  Chapter of the Association for Computational Linguistics: Human Language
  Technologies, Volume 1 (Long Papers)}, volume~1, pages 2227--2237.

\bibitem[{Pineau et~al.(2020)Pineau, Vincent-Lamarre, Sinha, Larivi{\`e}re,
  Beygelzimer, d'Alch{\'e} Buc, Fox, and Larochelle}]{pineau2020improving}
Joelle Pineau, Philippe Vincent-Lamarre, Koustuv Sinha, Vincent Larivi{\`e}re,
  Alina Beygelzimer, Florence d'Alch{\'e} Buc, Emily Fox, and Hugo Larochelle.
  2020.
\newblock Improving reproducibility in machine learning research (a report from
  the neurips 2019 reproducibility program).
\newblock \emph{arXiv preprint arXiv:2003.12206}.

\bibitem[{Qian et~al.(2018)Qian, Santus, Jin, Guo, and
  Barzilay}]{qian2018graphie:}
Yujie Qian, Enrico Santus, Zhijing Jin, Jiang Guo, and Regina Barzilay. 2018.
\newblock Graphie: A graph-based framework for information extraction.
\newblock \emph{arXiv: Computation and Language}.

\bibitem[{Sang and De~Meulder(2003)}]{sang2003introduction}
Erik~F Sang and Fien De~Meulder. 2003.
\newblock Introduction to the conll-2003 shared task: Language-independent
  named entity recognition.
\newblock \emph{arXiv preprint cs/0306050}.

\bibitem[{Schlichtkrull et~al.(2018)Schlichtkrull, Kipf, Bloem, Van Den~Berg,
  Titov, and Welling}]{schlichtkrull2018modeling}
Michael Schlichtkrull, Thomas~N Kipf, Peter Bloem, Rianne Van Den~Berg, Ivan
  Titov, and Max Welling. 2018.
\newblock Modeling relational data with graph convolutional networks.
\newblock In \emph{European Semantic Web Conference}, pages 593--607. Springer.

\bibitem[{Strubell et~al.(2017)Strubell, Verga, Belanger, and
  Mccallum}]{strubell2017fast}
Emma Strubell, Patrick Verga, David Belanger, and Andrew Mccallum. 2017.
\newblock Fast and accurate entity recognition with iterated dilated
  convolutions.
\newblock pages 2670--2680.

\bibitem[{Wadden et~al.(2019)Wadden, Wennberg, Luan, and
  Hajishirzi}]{wadden2019entity}
David Wadden, Ulme Wennberg, Yi~Luan, and Hannaneh Hajishirzi. 2019.
\newblock Entity, relation, and event extraction with contextualized span
  representations.
\newblock \emph{arXiv preprint arXiv:1909.03546}.

\bibitem[{Wilcoxon et~al.(1970)Wilcoxon, Katti, and
  Wilcox}]{wilcoxon1970critical}
Frank Wilcoxon, SK~Katti, and Roberta~A Wilcox. 1970.
\newblock Critical values and probability levels for the wilcoxon rank sum test
  and the wilcoxon signed rank test.
\newblock \emph{Selected tables in mathematical statistics}, 1:171--259.

\bibitem[{Xu et~al.(2019)Xu, Yang, Kang, Wang, and Liu}]{xu2019document}
Kai Xu, Zhenguo Yang, Peipei Kang, Qi~Wang, and Wenyin Liu. 2019.
\newblock Document-level attention-based bilstm-crf incorporating disease
  dictionary for disease named entity recognition.
\newblock \emph{Computers in biology and medicine}, 108:122--132.

\bibitem[{Yan et~al.(2019)Yan, Deng, Li, and Qiu}]{yan2019tener}
Hang Yan, Bocao Deng, Xiaonan Li, and Xipeng Qiu. 2019.
\newblock Tener: Adapting transformer encoder for name entity recognition.
\newblock \emph{arXiv preprint arXiv:1911.04474}.

\bibitem[{Yang et~al.(2018)Yang, Liang, and Zhang}]{yang2018design}
Jie Yang, Shuailong Liang, and Yue Zhang. 2018.
\newblock Design challenges and misconceptions in neural sequence labeling.
\newblock \emph{arXiv: Computation and Language}.

\bibitem[{Yang et~al.(2019)Yang, Dai, Yang, Carbonell, Salakhutdinov, and
  Le}]{yang2019xlnet}
Zhilin Yang, Zihang Dai, Yiming Yang, Jaime Carbonell, Russ~R Salakhutdinov,
  and Quoc~V Le. 2019.
\newblock Xlnet: Generalized autoregressive pretraining for language
  understanding.
\newblock In \emph{Advances in neural information processing systems}, pages
  5753--5763.

\bibitem[{Ye et~al.(2021)Ye, Liu, Fu, and Neubig}]{ye2021towards}
Zihuiwen Ye, Pengfei Liu, Jinlan Fu, and Graham Neubig. 2021.
\newblock Towards more fine-grained and reliable nlp performance prediction.
\newblock \emph{arXiv preprint arXiv:2102.05486}.

\bibitem[{Yu et~al.(2020)Yu, Jiang, Yang, and Xia}]{yu2020improving}
Jianfei Yu, Jing Jiang, Li~Yang, and Rui Xia. 2020.
\newblock Improving multimodal named entity recognition via entity span
  detection with unified multimodal transformer.
\newblock Association for Computational Linguistics.

\end{thebibliography}
\bibliographystyle{acl_natbib}

\appendix

\renewcommand\tabcolsep{2pt}
\begin{table*}[!ht]
  \centering \footnotesize
    \begin{tabular}{cccccccccccccc}
    \toprule
    \textbf{Agg.}    & \texttt{CITYU} & \texttt{NCC}   & \texttt{SXU}   & \texttt{PKU}   & \texttt{CN03}  & \texttt{BC}    & \texttt{BN}    & \texttt{MZ}    & \texttt{WB}    & \texttt{NW}    & \texttt{TC}    & \texttt{CN00}  & \texttt{PTB} \\
    \midrule
    \textit{bow}       & 5     & 6     & 2     & 2     & 2     & 5     & 4     & 2     & 10    & 2     & 5     & 2     & 4 \\
    \textit{graph}     & 7     & 3     & 3     & 6     & 10    & 6     & 4     & 3     & 7     & 6     & 10    & 3     & 10 \\
    \textit{seq}     & 7     & 3     & 3     & 6     & 10    & 6     & 4     & 3     & 7     & 6     & 10    & 3     & 10 \\
    \textit{cPre}     & 7     & 3     & 3     & 6     & 10    & 7     & 4     & 3     & 7     & 6     & 10    & 2     & 10 \\
    \bottomrule
    \end{tabular}%
    \caption{The window size $k$ when the four larger-context aggregators  achieve the final performance. }
  \label{tab:window-size}%
\end{table*}%

\renewcommand\tabcolsep{2.5pt}
\begin{table*}[htb]
  \centering  
    \begin{tabular}{lccccccccc}
    \toprule
    \textbf{Task} & \textbf{Data} & \texttt{eCon} & \texttt{tCon} & \texttt{eFre} & \texttt{tFre} & \texttt{eLen} & \texttt{dOov} & \texttt{sLen} & \texttt{eDen} \\
    \midrule
    \multirow{7}[1]{*}{NER} & CN03 & 0.485  & 0.514  & 0.109  & 0.017  & 1.436  & 0.067  & 13.4  & 0.232  \\
          & BC & 0.486  & 0.440  & 0.113  & 0.108  & 1.905  & 0.064  & 16.3  & 0.085  \\
          & BN & 0.627  & 0.552  & 0.147  & 0.089  & 1.623  & 0.004  & 19.5  & 0.148  \\
          & MZ & 0.496  & 0.487  & 0.129  & 0.119  & 1.832  & 0.015  & 22.9  & 0.124  \\
          & WB & 0.294  & 0.269  & 0.111  & 0.084  & 1.631  & 0.024  & 22.9  & 0.050  \\
          & NW & 0.567  & 0.512  & 0.083  & 0.108  & 2.015  & 0.014  & 26.1  & 0.179  \\
          & TC & 0.261  & 0.258  & 0.082  & 0.105  & 1.598  & 0.043  & 8.3   & 0.040  \\
    \bottomrule
    \toprule
    \textbf{Task} & \textbf{Data} & \texttt{wCon} & \texttt{cCon} & \texttt{wFre} & \texttt{cFre} & \texttt{wLen} & \texttt{dOov} & \texttt{sLen} &  \\
    \midrule
    \multirow{4}[1]{*}{\textbf{CWS}} 
          & CITYU & 0.763  & 0.285  & 1.834  & 0.489  & 1.634  & 0.010  & 62.400  &  \\
          & NCC & 0.743  & 0.274  & 3.856  & 1.016  & 1.546  & 0.021  & 64.400  &  \\
          & SXU & 0.781  & 0.293  & 3.832  & 1.005  & 1.590  & 0.020  & 69.700  &  \\
          & PKU & 0.777  & 0.292  & 1.765  & 0.466  & 1.615  & 0.019  & 59.200  &  \\
    \bottomrule
    \end{tabular}%
  \caption{The data-level measure $\zeta_p$ in seven (four) datasets with respect to eight (seven) attributes in NER (CWS) task. 
The value of \textit{wFre} and \textit{cFre} on CWS task needs to multiply by $10^{-7}$.}
  \label{tab:data-measure}%
\end{table*}%

\section{Aggregator Setting}
Tab.~\ref{tab:window-size} illustrates the window size $k (k \neq 1)$ when the larger-context aggregator achieves the best performance. The window size $k$ when \textit{seq} achieves the best performance will be chosen to set the document-length of the \textit{graph} aggregator.

\renewcommand\tabcolsep{1.2pt}
\begin{table*}[htb]
  \centering \footnotesize
    \begin{tabular}{p{0.4cm}lcccccccc}
    \toprule
         & \textbf{Attr.} & \texttt{eCon} & \texttt{tCon} & \texttt{eFre} & \texttt{tFre} & \texttt{eLen} & \texttt{dOov} & \texttt{sLen} & \texttt{eDen} \\
    \midrule
    \multicolumn{2}{c}{\multirow{4}[2]{*}{\textbf{$ \zeta_p$}}} & \multirow{4}[2]{*}{\includegraphics[scale=0.13]{pic/radar/ner_eCon.pdf}} & 
    \multirow{4}[2]{*}{\includegraphics[scale=0.13]{pic/radar/ner_tCon.pdf}} & 
    \multirow{4}[2]{*}{\includegraphics[scale=0.13]{pic/radar/ner_eFre.pdf}} & 
    \multirow{4}[2]{*}{\includegraphics[scale=0.13]{pic/radar/ner_tFre.pdf}} & 
    \multirow{4}[2]{*}{\includegraphics[scale=0.13]{pic/radar/ner_eLen.pdf}} & 
    \multirow{4}[2]{*}{\includegraphics[scale=0.13]{pic/radar/ner_dOov.pdf}} & 
    \multirow{4}[2]{*}{\includegraphics[scale=0.13]{pic/radar/ner_sLen.pdf}} & 
    \multirow{4}[2]{*}{\includegraphics[scale=0.13]{pic/radar/ner_eDen.pdf}} \\
         \\  \\ \\ 
    \midrule
    \multirow{4}[2]{*}{\textbf{$\rho$}} & \textit{bow} & \textcolor{cadetgrey}{-0.179}  & -0.607  & \textcolor{cadetgrey}{0.143}  & 0.396  & 0.643  & \textcolor{cadetgrey}{-0.036}  & \textcolor[rgb]{0.0, 0.5, 0.0}{0.468}  & -0.571  \\
    & \textit{graph} & -0.571  & \textcolor{cadetgrey}{-0.143}  & -0.393  & \textcolor[rgb]{0.0, 0.5, 0.0}{-0.919}  & -0.643  & 0.286  & -0.288  & \textcolor{cadetgrey}{-0.107}  \\
    & \textit{seq} & -0.643  & \textcolor[rgb]{0.0, 0.5, 0.0}{-0.857}  & -0.429  & \textcolor{cadetgrey}{0.162}  & \textcolor{cadetgrey}{0.143}  & \textcolor{cadetgrey}{0.071}  & \textcolor{cadetgrey}{0.180}  & \textcolor[rgb]{0.0, 0.5, 0.0}{-0.714}  \\
    & \textit{cPre} & \textcolor[rgb]{0.0, 0.5, 0.0}{-0.714}  & -0.750  & \textcolor[rgb]{0.0, 0.5, 0.0}{-0.571}  & -0.306  & \textcolor{cadetgrey}{0.000}  & \textcolor[rgb]{0.0, 0.5, 0.0}{0.357}  & \textcolor{cadetgrey}{-0.180}  & -0.643  \\
    \bottomrule
    \end{tabular}%
 \caption{Illustration of measures $\zeta_p$ in seven datasets (\texttt{CN03}, \texttt{TC}, \texttt{NW}, \texttt{WB}, \texttt{MZ}, \texttt{BN}, \texttt{BC}) with respect to eight attributes (e.g., \texttt{eCon}) and correlation measure $\rho$ in NER task. 
 A higher absolute value $\rho$ (e.g \texttt{|-0.714|})  represents the improvement of corresponding aggregator (e.g., \texttt{seq}) heavily correlate with corresponding attribute (e.g. \texttt{eCon}). 
 The number with the highest absolute value of each column is colored by green.
 ``cPre'' represents ``cPre-seq'' and the value in grey denotes correlation value does not pass a significance test ($p=0.05$).  
 }
    \label{tab:spearman-data-metric2}%
\end{table*}%

\section{Quantifying and Understanding Dataset Bias}
In this section, we will supplement some analyses related to Sec.~\ref{sec:exp-I}. 

\subsection{Data-level Measure}
Tab.~\ref{tab:data-measure} gives the data-level measure $\zeta_p$ in seven (four) datasets with respect to eight (seven) attributes in NER (CWS) task. The data-level measure $\zeta_p$ will be used to compute the \textit{correlation measure} in Sec.~\ref{sec:exp-I}.

\subsection{Results}

Tab.~\ref{tab:spearman-data-metric2} 
displays (using spider charts) measure $\zeta_p$ of seven datasets with respect to diverse attributes, and correlation measure $\rho$ in NER task. We have given a detail analysis on \textit{seq} and \textit{cPre-seq} on the main text, here, we will provide the suggestion for choosing the datasets for \textit{bow} and \textit{graph} aggregator.

\noindent (1) regarding \textit{bow} aggregator, it negatively correlated with $\zeta_{\texttt{tCon}}$  and $\zeta_{\texttt{eDen}}$ with larger correlation values. Therefore, \textit{bow} aggregator is more appropriate to deal with datasets \texttt{WB}, \texttt{TC}, \texttt{BC},
since these four datasets are with lower value of $\zeta_p$ with respect to the attribute  \texttt{tCon} (\texttt{TC}, \texttt{WB}) and \texttt{eDen} (\texttt{BC}, \texttt{WB}, \texttt{TC}). 
Additionally, \texttt{bow} aggregator obtained the highest positive correlation with $\zeta_{\texttt{tFre}}$, $\zeta_{\texttt{eLen}}$, and $\zeta_{\texttt{sLen}}$. Besides,
$\zeta_{\texttt{tFre}}(\texttt{MZ})$,
$\zeta_{\texttt{eLen}}(\texttt{NW})$, and
$\zeta_{\texttt{sLen}}(\texttt{NW})$, also achieved the highest value, suggesting that \texttt{bow} aggregator is suitable for \texttt{MZ} and \texttt{NW}.

\noindent (2) regarding \textit{graph} aggregator, it negatively correlated with $\zeta_{\texttt{eCon}}$, $\zeta_{\texttt{eFre}}$, $\zeta_{\texttt{tFre}}$, and $\zeta_{\texttt{eLen}}$,  with larger correlation values. Therefore, \texttt{graph} aggregator is more appropriate to deal with datasets \texttt{WB}, \texttt{TC}, \texttt{NW}, and \texttt{CN03},
since these four datasets are with lower value of $\zeta_p$ with respect to the attribute 
\texttt{eCon} (\texttt{TC,WB}),  
\texttt{eFre} (\texttt{NW}, \texttt{TC}),  
\texttt{tFre} (\texttt{CN03}, \texttt{WB}), and 
\texttt{eLen} (\texttt{CN03}),.

\renewcommand\tabcolsep{1.5pt}
\begin{table*}[htb]
  \centering \footnotesize
    \begin{tabular}{p{0.4cm}lccccccc}
    \toprule
         & \textbf{Attr.} & \texttt{wCon} & \texttt{cCon} & \texttt{wFre} & \texttt{cFre} & \texttt{wLen} & \texttt{dOov} & \texttt{sLen} \\
    \midrule
    \multicolumn{2}{c}{\multirow{4}[2]{*}{\textbf{$ \zeta_p$}}} & \multirow{4}[2]{*}{\includegraphics[scale=0.13]{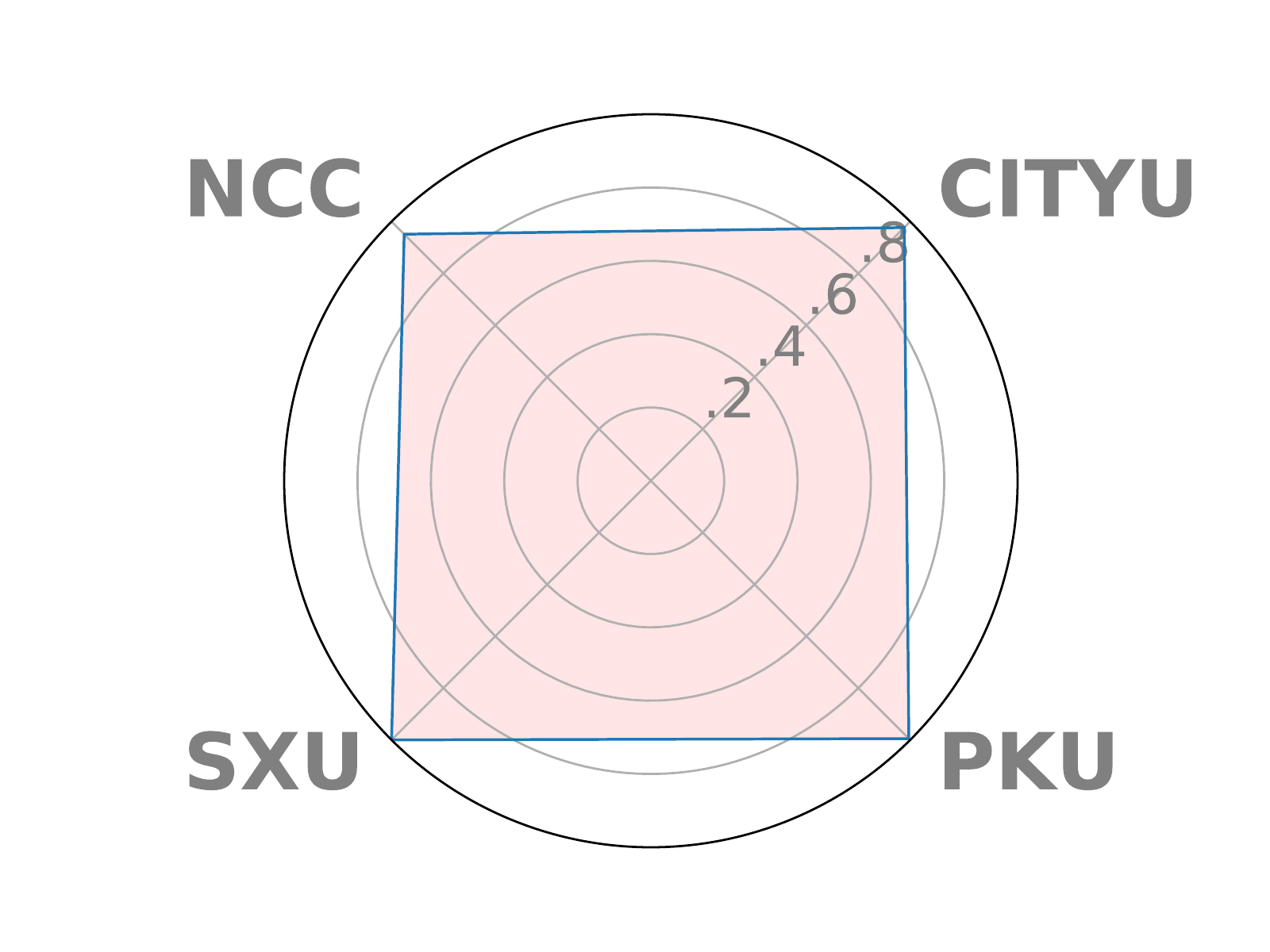}} & 
    \multirow{4}[2]{*}{\includegraphics[scale=0.13]{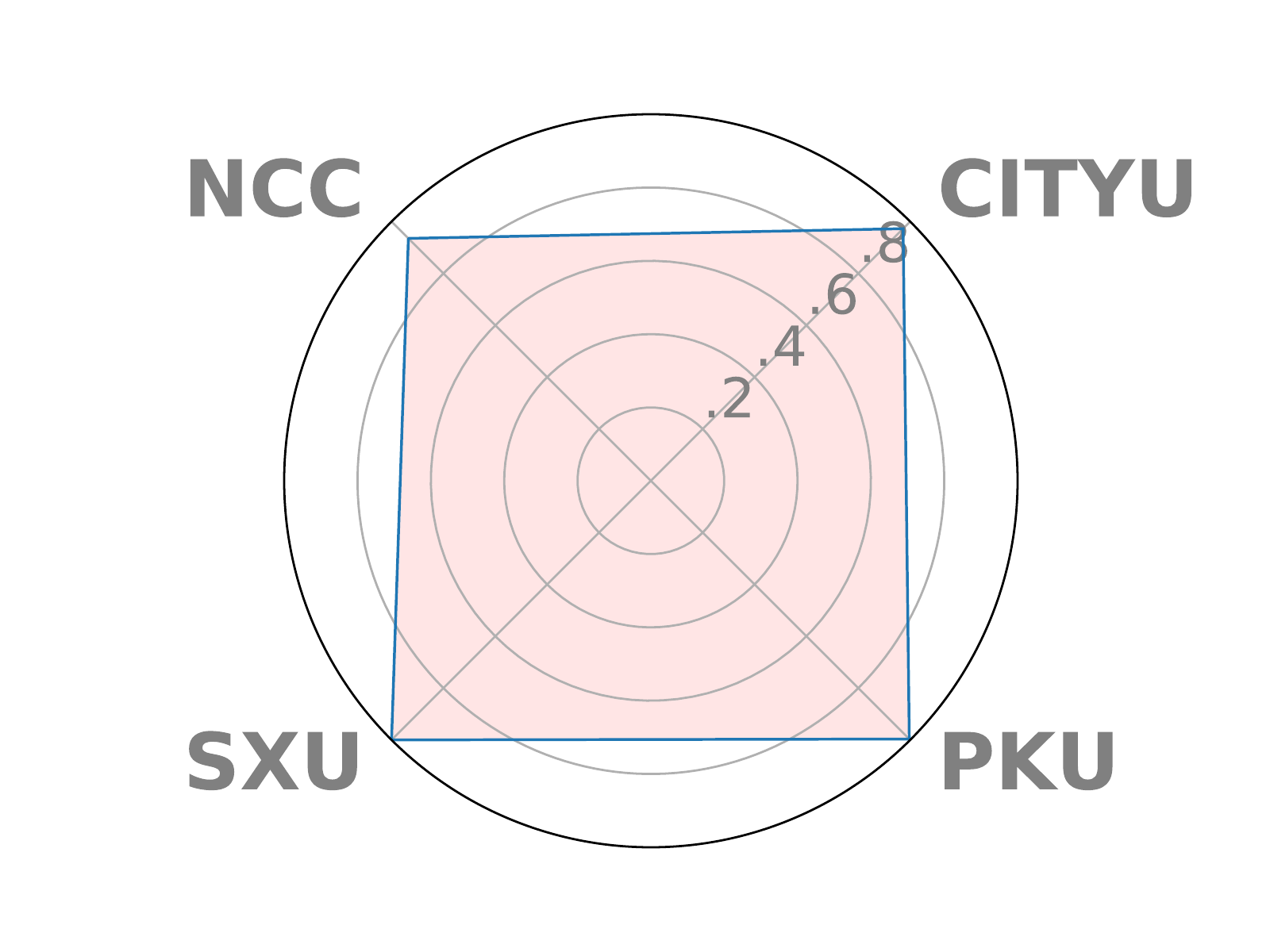}} & 
    \multirow{4}[2]{*}{\includegraphics[scale=0.13]{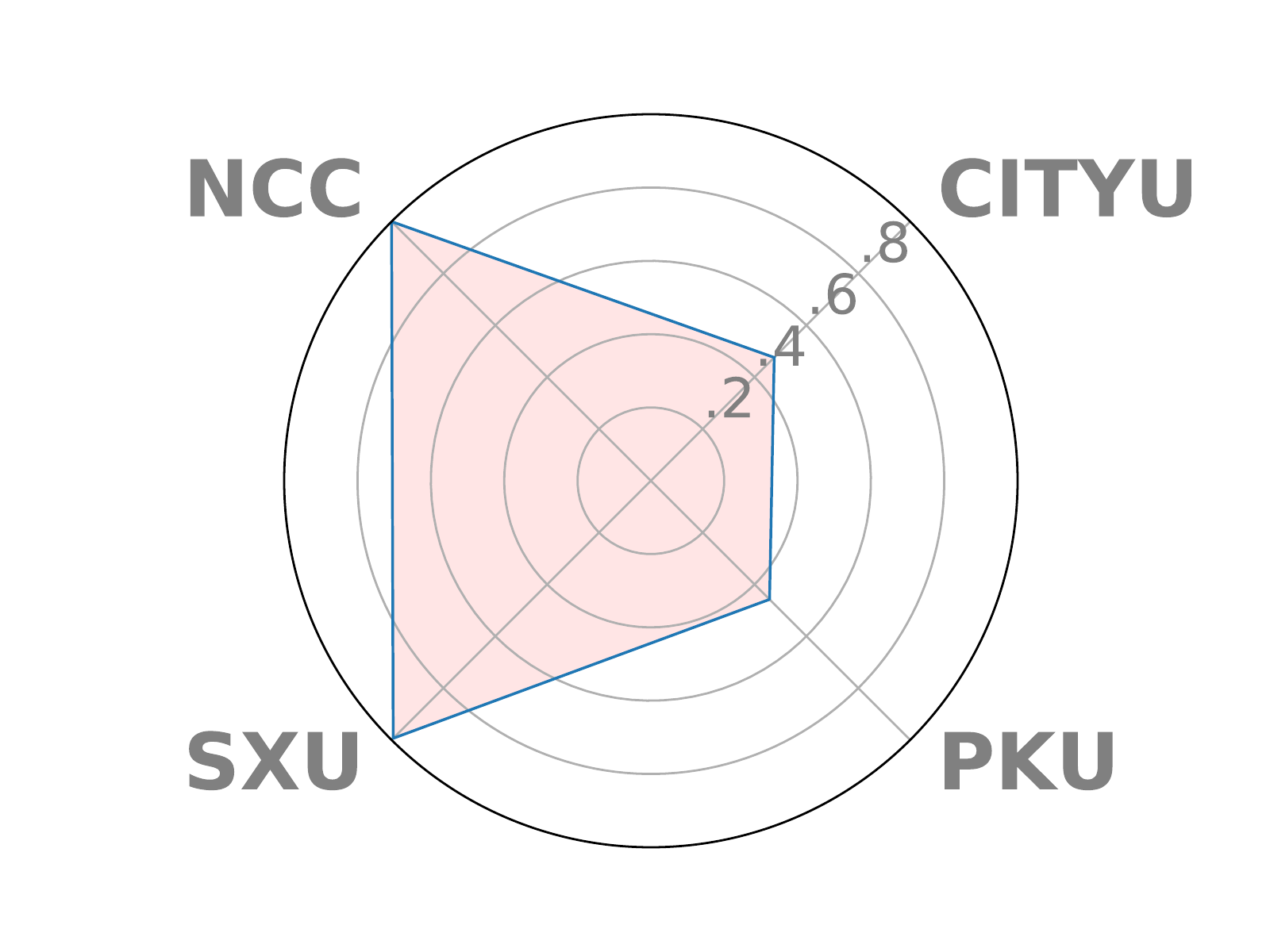}} & 
    \multirow{4}[2]{*}{\includegraphics[scale=0.13]{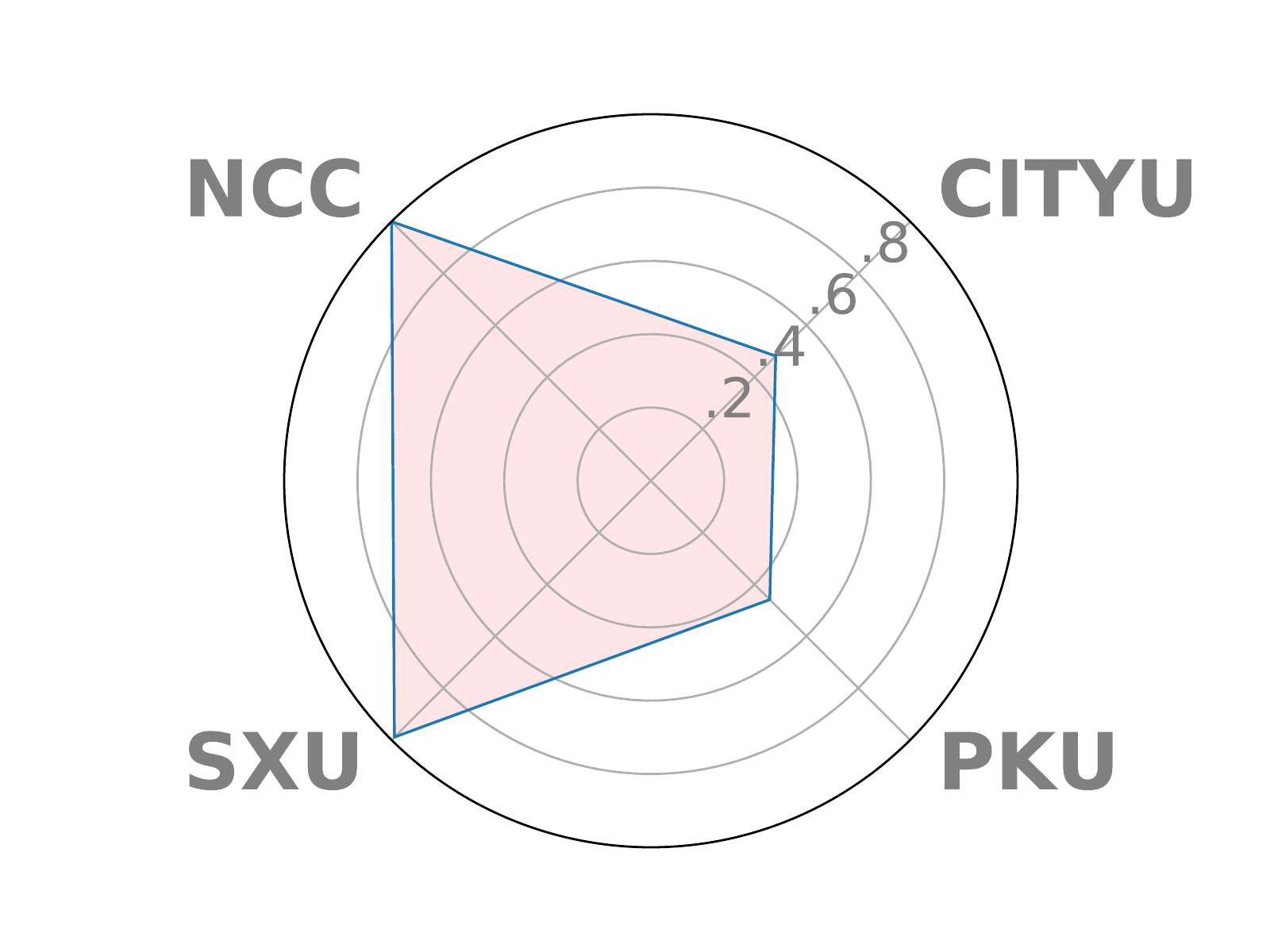}} & 
    \multirow{4}[2]{*}{\includegraphics[scale=0.13]{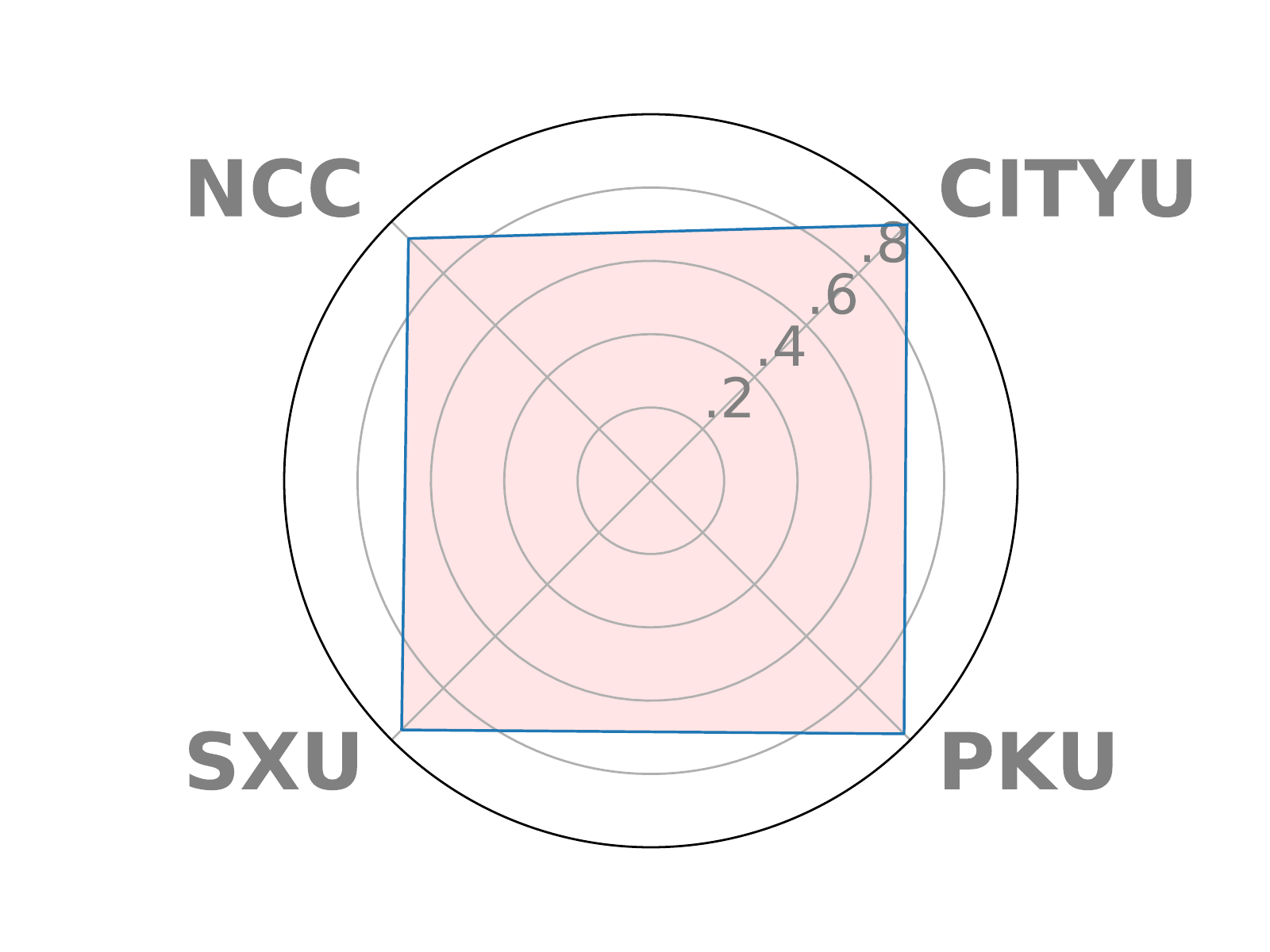}} & 
    \multirow{4}[2]{*}{\includegraphics[scale=0.13]{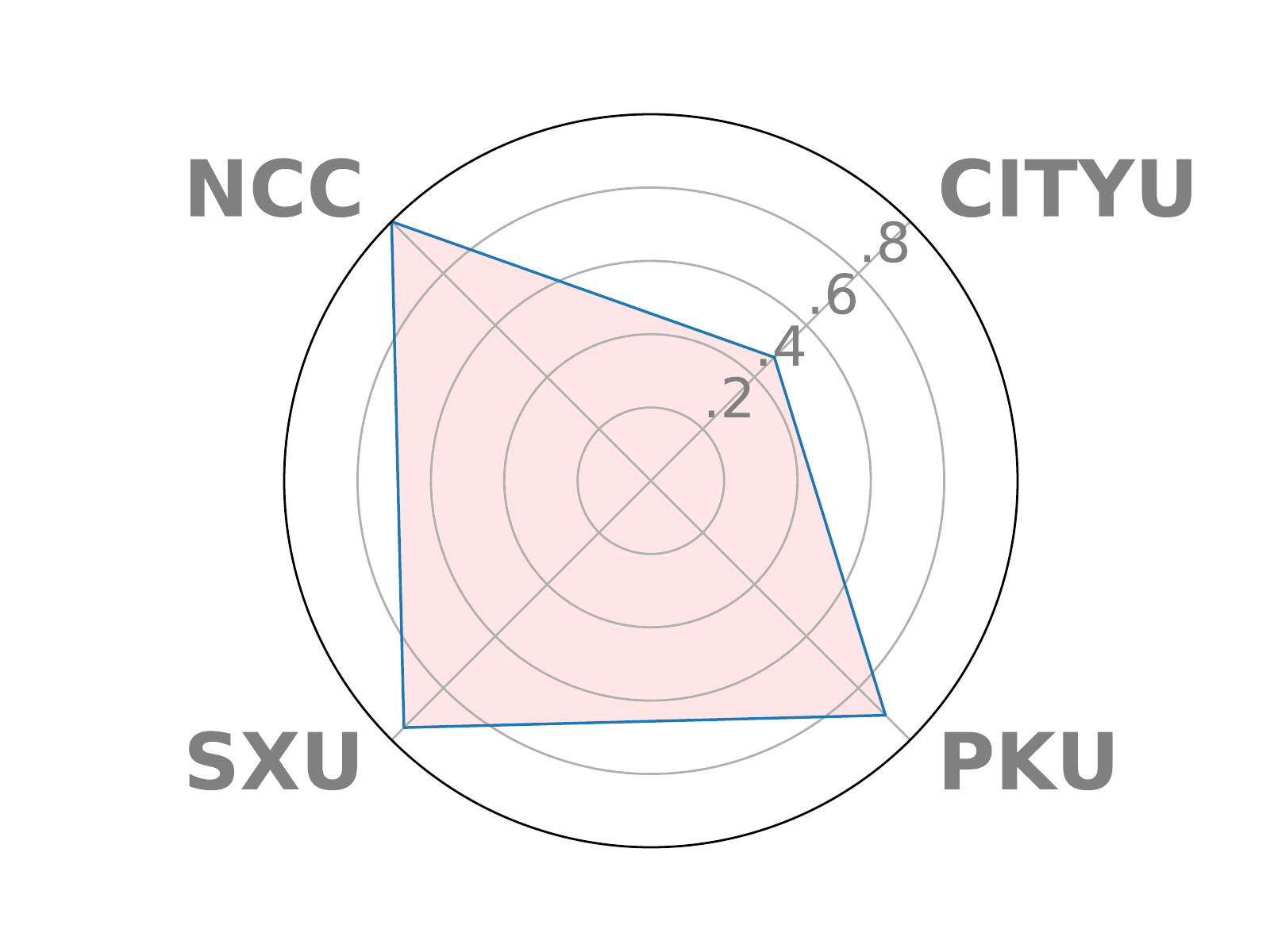}} & 
    \multirow{4}[2]{*}{\includegraphics[scale=0.13]{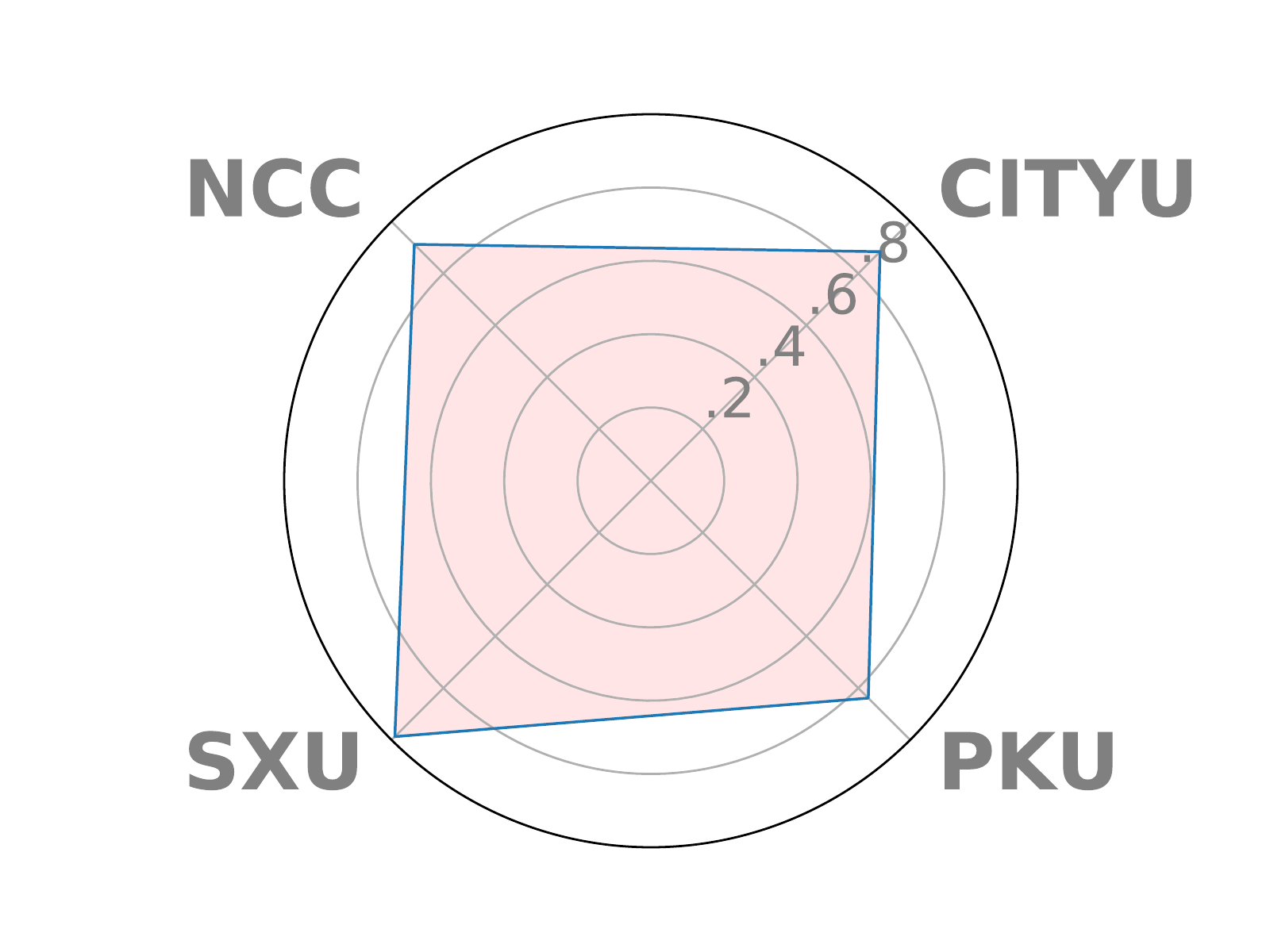}} \\
         \\  \\ \\ 
    \midrule
    \multirow{4}[2]{*}{\textbf{$\rho$}} & \textit{bow}  & \textcolor[rgb]{0.0, 0.5, 0.0}{-0.657}  & \textcolor[rgb]{0.0, 0.5, 0.0}{-0.771}  & \textcolor{cadetgrey}{0.086}  & 0.257  & \textcolor[rgb]{0.0, 0.5, 0.0}{-0.600}  & 0.319  & \textcolor[rgb]{0.0, 0.5, 0.0}{-0.486} \\
    & \textit{graph} & \textcolor{cadetgrey}{-0.029}  & 0.257  & \textcolor[rgb]{0.0, 0.5, 0.0}{-0.543}  & \textcolor[rgb]{0.0, 0.5, 0.0}{-0.429}  & \textcolor{cadetgrey}{0.143}  & -0.319  & \textcolor[rgb]{0.0, 0.5, 0.0}{-0.486}  \\
    & \textit{seq} & \textcolor{cadetgrey}{0.086}  & \textcolor{cadetgrey}{-0.200}  & 0.371  & 0.314  & -0.257  & 0.464  & 0.257  \\
    & \textit{cPre} & 0.580  & 0.493  & -0.261  & \textcolor{cadetgrey}{-0.203}  & \textcolor{cadetgrey}{-0.203}  & \textcolor[rgb]{0.0, 0.5, 0.0}{0.544}  & \textcolor{cadetgrey}{-0.232}  \\
    \bottomrule
    \end{tabular}%
 \caption{Illustration of measures $\zeta_p$ in four datasets ( \texttt{CITYU}, \texttt{NCC}, \texttt{SXU}, \texttt{PKU}) with respect to seven attributes (e.g., \texttt{wCon}) and correlation measure $\rho$ in CWS task. 
 A higher absolute value $\rho$ (e.g \texttt{|-0.657|})  represents the improvement of corresponding aggregator (e.g., \texttt{bow}) heavily correlate with corresponding attribute (e.g. \texttt{wCon}). 
 The number with the highest absolute value of each column is colored by green.
 ``cPre'' represents ``cPre-seq'' and the value in grey denotes correlation value does not pass a significance test ($p=0.05$). 
 }
    \label{tab:cws-spearman-data-metric}%
\end{table*}%

Tab.~\ref{tab:cws-spearman-data-metric} illustrates the measures $\zeta_p$ in four CWS datasets with respect to seven attributes (e.g., \texttt{wCon}) and correlation measure $\rho$. 
We can conduct similar analysis like NER for CWS. We highlight the suitable  CWS datasets for each aggregator as follows: 

\begin{itemize}
    \item \textit{bow}: \texttt{NCC}, and \texttt{SXU}.
    \vspace{-2pt}
    \item \textit{graph}:  \texttt{PKU} and \texttt{CITYU}.  
    \vspace{-2pt}
    \item \textit{seq}: \texttt{SXU}, and \texttt{NCC}.
    \vspace{-2pt}
    \item \textit{cPre-seq}: \texttt{CITYU}, \texttt{PKU}, \texttt{SXU}.
    \vspace{-2pt}
\end{itemize}

\end{document}